\documentclass[journal]{IEEEtran}

\usepackage[noadjust]{cite}
\usepackage{amsmath}
\usepackage{url}
\usepackage{xcolor}
\usepackage{times}
\usepackage{epsfig}
\usepackage{graphicx}
\usepackage{caption}
\usepackage{float}
\usepackage{amssymb} 
\usepackage{footmisc}
\usepackage{lineno}
\usepackage{color}
\usepackage{subfigure}
\usepackage{multirow}
\usepackage{dsfont}
\usepackage{mathtools}
\usepackage{setspace}
\usepackage{csquotes}
\usepackage{breqn}
\usepackage{lettrine}
\usepackage{upquote}
\usepackage{adjustbox}
\usepackage{wrapfig}
\usepackage{lipsum}
\usepackage{mathtools}
\usepackage{comment}
\usepackage{bm}
\usepackage{array}
\usepackage{tabu}
\usepackage{booktabs}
\usepackage{enumerate}
\DeclareCaptionFormat{myformat}{\fontsize{8}{10}\selectfont#1#2#3}
\DeclareMathOperator*{\argmin}{argmin}
\DeclareMathAlphabet\mathbfcal{OMS}{cmsy}{b}{n}

\DeclareMathAlphabet{\pazocal}{OMS}{zplm}{m}{n}
\DeclareMathAlphabet{\mathpzc}{OT1}{pzc}{m}{it}

\usepackage[ruled,vlined]{algorithm2e}
\usepackage[colorlinks=true,linkcolor=blue]{hyperref}
\def\arrvline{\hfil\kern\arraycolsep\vline\kern-\arraycolsep\hfilneg}

\makeatletter
\newenvironment{subxarray}{%
  \vcenter\bgroup
  \Let@ \restore@math@cr \default@tag
  \baselineskip\fontdimen10 \scriptfont\tw@
  \advance\baselineskip\fontdimen12 \scriptfont\tw@
  \lineskip\thr@@\fontdimen8 \scriptfont\thr@@
  \lineskiplimit\lineskip
  \ialign\bgroup\hfil
    $\m@th\scriptstyle##$&$\m@th\scriptstyle{}##$\hfil\crcr
}{%
  \crcr\egroup\egroup
}
\makeatother

\ifCLASSINFOpdf
\else
\fi

\begin{document}
\bstctlcite{IEEEexample:BSTcontrol}
\title{Learning Spatial-Temporal Regularized Tensor Sparse RPCA for Background Subtraction}

\author{Basit Alawode and Sajid Javed 
\thanks{B. Alawode and S. Javed are with the department of electrical engineering and computer science, Khalifa University of Science and Technology, P.O Box: 127788, Abu Dhabi, UAE. (email: sajid.javed@ku.ac.ae).}
}
\maketitle

\begin{abstract}
Video background subtraction is one of the fundamental problems in computer vision that aims to segment all moving objects.
Robust Principal Component Analysis (RPCA) has been identified as a promising unsupervised paradigm for background subtraction tasks in the last decade thanks to its competitive performance in a number of benchmark datasets.  
Tensor RPCA (TRPCA) variations have improved background subtraction performance further. 
However, because moving object pixels in the sparse component are treated independently and don't have to adhere to spatial-temporal structured-sparsity constraints, performance is reduced for sequences with dynamic backgrounds, camouflaged, and camera jitter problems.  
In this work, we present a spatial-temporal regularized tensor sparse RPCA algorithm for precise background subtraction. 
Within the sparse component, we impose spatial-temporal regularizations in the form of normalized graph-Laplacian matrices.  
To do this, we build two graphs, one across the input tensor's spatial locations and the other across its frontal slices in the time domain.  
While maximizing the objective function, we compel the tensor sparse component to serve as the spatiotemporal eigenvectors of the graph-Laplacian matrices.  
The disconnected moving object pixels in the sparse component are preserved by the proposed graph-based regularizations since they both comprise of spatiotemporal subspace-based structure. 
Additionally, we propose a unique objective function that employs batch and online-based optimization methods to jointly maximize the background-foreground and spatial-temporal regularization components. 
Experiments are performed on six publicly available background subtraction datasets that demonstrate the superior performance of the proposed algorithm compared to several existing methods. 
Our source code will be available very soon.
\end{abstract}

\IEEEpeerreviewmaketitle

\section{Introduction}
\noindent \lettrine[lraise=0.1, nindent=0em, slope=-.5em]{V}ideo background subtraction also known as moving object segmentation from static camera is one of the long-standing problems in computer vision \cite{garcia2020background, bouwmans2019deep, bouwmans2017decomposition}. 
 The primary goal of background subtraction is to separate moving objects from the background model, a static scene \cite{garcia2020background}.
Background subtraction has numerous applications including video surveillance \cite{brutzer2011evaluation}, semantic segmentation \cite{braham2017semantic, long2015fully}, object detection and tracking \cite{javed2022visual, elhoseny2020multi}, autonomous driving \cite{muro2021moving}, robotics manipulation \cite{ainetter2021end}, sports video analysis \cite{d2010review}, and human activity recognition \cite{aggarwal2011human}.
However, it becomes extremely difficult when there are dynamic backgrounds present, such as swaying bushes, rippling water, varying lighting, irregular object motion, bad weather, camouflaged foreground objects, pan-tilt-zoom camera sequences, and extreme nighttime scenes \cite{garcia2020background, bouwmans2019deep, zhang2021meta, zhao2022universal, zheng2020novel, tezcan2020bsuv}. 
Numerous approaches have been developed in the literature to solve the aforementioned problems, including statistical background modeling \cite{garcia2020background, bouwmans2011recent}, subspace learning models \cite{bouwmans2017decomposition}, and deep learning models \cite{bouwmans2019deep}. 
Background subtraction is still a difficult challenge for scenes with varying backgrounds and shadows, though \cite{tezcan2020bsuv, zhao2022universal, li2022tensor, li2018online, sultana2022unsupervised}.

\begin{figure}[t!]
\centering
\includegraphics[width=\linewidth]{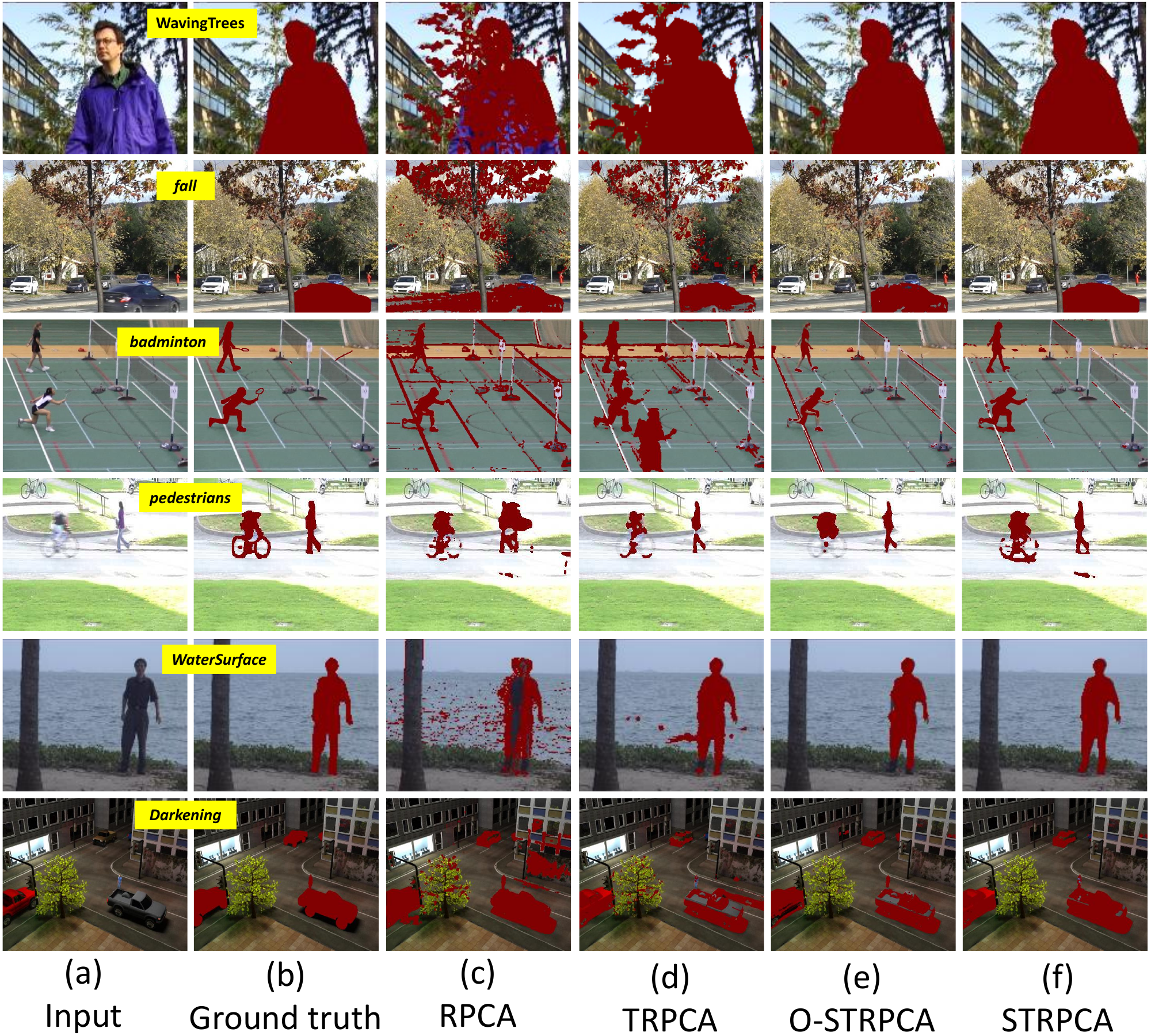}
\caption{Results of background subtraction using RPCA, TRPCA, and our proposed algorithms on a number of difficult sequences chosen from openly accessible datasets.
From left to right, (a) displays sample input images, (b) displays ground-truth images, (c) displays background subtraction results using RPCA \cite{candes2011robust}, (d) displays results using TRPCA \cite{lu2019tensor}, and (e)–(f) displays results estimated using our proposed O-STRPCA and STRPCA algorithms.
Selected sequences from the CD14, I2R, SABS, Wallflower, and I2R datasets are displayed, going from top to bottom.}
\label{fig1}
\end{figure}

Robust Principal Component Analysis (RPCA) and its variant Tensor RPCA (TRPCA) are popular unsupervised paradigms and have been successfully used in many problems \cite{bouwmans2017decomposition, vaswani2018robust, bouwmans2018applications}.
This has included background-foreground separation problems \cite{bouwmans2017decomposition}, salient object detection \cite{peng2016salient}, image or video denoising \cite{wen2020image, xue2021multilayer}, data clustering \cite{yin2018subspace}, medical applications \cite{li2019specular}, and hyperspectral imaging  \cite{zhai2018laplacian, wang2021tensor} in the past decade.
Wright \textit{et. al} posed the background subtraction problem as an RPCA-based matrix decomposition problem into the sum of its low-rank and sparse components \cite{wright2009robust}.
The temporal background sequence is highly correlated; therefore, the background model is located in a low-dimensional redundant subspace known as the low-rank background component. 
The grossly corrupted sparse component is made up of locally distinct regions known as foreground segmentation. 
By employing the Alternating Direction Method of Multipliers (ADMM) to solve the convex optimization problem, such a matrix decomposition may be produced \cite{boyd2011distributed}. 

RPCA has shown good performance for background subtraction \cite{bouwmans2017decomposition}. 
It is only useful for processing 2D matices, though. 
Instead of a 2D matrix, real-time sequences may contain multi-dimensional data that generates a 3D matrix. 
Therefore, before estimating background-foreground components using RPCA, a reshaping step is typically required. 
Such a procedure would potentially damage the sequence's inherent spatial patterns, which would reduce performance.
Several variations that enforce structural restrictions are suggested in the literature to enhance background subtraction performance, although all methods need a reshaping phase \cite{zhou2012moving, cao2015total, ebadi2017foreground, javed2016spatiotemporal, javed2017background, javed2018moving}.
In situations involving dense moving objects and really small moving objects, the moving object region therefore also becomes over-smoothed (Fig. \ref{fig1} (c)).

TRPCA approaches, which expand matrix-based RPCA and take advantage of the intrinsic multi-dimensional structure of input data matrix, have recently been developed to solve this deficiency \cite{lu2016tensor, lu2019tensor, goldfarb2014robust}.
Lu \textit{et al.} formulated the tensor-based decomposition framework as follows: \cite{lu2019tensor}:

\begin{equation}
\min_{\mathbfcal{B},\mathbfcal{F}}||\mathbfcal{B}||_{*}+\lambda||\mathbfcal{F}||_{1},\textrm{ such that}~\mathbfcal{X}=\mathbfcal{B}+\mathbfcal{F},
\label{eqn1}
\end{equation}

\noindent where $\mathbfcal{X} \in \mathbb{R}^{w \times h \times n}$ is the input tensor and each $i$-th frame in this tensor is denoted by $\textbf{X}_{i}\in \mathbb{R}^{w \times h}$ having width $w$ and height $h$, respectively.
$\lambda=1 / \sqrt{\max (w,h,n)}$ assigns relative importance while optimizing (\ref{eqn1}).
Model (\ref{eqn1}) perfectly extracts the low-rank tensor $\mathbfcal{B}$ comprising the background model and the sparse tensor $\mathbfcal{F}$ constituting moving object segmentation under specific incoherence conditions.
The segmentation of moving objects can be done more effectively with TRPCA and its variants \cite{shakeri2019moving, hu2016moving}.
There are still two issues that must be resolved, though.
The $\ell_{1}$-norm regularization on $\mathbfcal{F}$ handles each pixel independently and therefore ignores the spatial coherent structure. 
This results in a performance drop for scenes with dynamic backgrounds (Fig. \ref{fig1} (c)-(d)). 
The $\mathbfcal{F}$ has a homogeneous spatial structure and may be handled cogently.
Therefore, the importance of fostering organized sparsity inside the sparse tensor $\mathbfcal{F}$, as noted, remains an unresolved problem; sadly, relatively few attempts have been made in this regard \cite{sobral2017matrix}.
Batch optimization of the model (\ref{eqn1}) requires all video frames must be present in memory before any processing.
Real-time processing is therefore compromised, which is necessary for surveillance videos.

In the current work, we address the aforementioned challenges by proposing a novel algorithm known as the Spatial-temporal regularized TRPCA (STRPCA) for background subtraction.
To lessen the effects of inaccurate pixels in the $\mathbfcal{F}$ component, we maintain the spatial and temporal structures of the moving object in the proposed algorithm.
We build two graphs—one spatial and the other temporal—for this aim.
The spatial-temporal structure of the $\mathbfcal{F}$ component is influenced by both graphs.
For each frontal slice of $\mathbfcal{X}$, a pixel-wise spatial graph is constructed using the nearest neighbor method. 
The spatial graph, in particular, imposes every pixel of the moving object to share a value with its linked neighbors.
To temporally constrain each pixel of the moving object to have a comparable value, a temporal graph is built among the frontal slices of the tensor $\mathbfcal{X}$.
To capture the notion of pairwise similarities inside the model (\ref{eqn1}), we estimate the normalized spatial and temporal graph-Laplacian matrices from both graphs \cite{yin2015laplacian}.
By requiring a model (\ref{eqn1}) to act as the eigenvectors of these matrices, one can be sure that the resulting model will capture the coherent and accurate structure of moving objects in $\mathbfcal{F}$.
This is because the eigenvectors of the corresponding Laplacian matrices preserve the spatial-temporal structure. 
By enforcing the STRPCA model to be the eigenvectors of the spatial-temporal Laplacian matrices, we compel it to be aware of both spatial and temporal $\mathbfcal{F}$ structure.
Our proposed algorithm is able to better discriminate the moving objects from their background even in the presence of camouflage, shadows, and dynamic backgrounds and thus improve the background subtraction performance by encoding these spectral clustering-based constraints into the TRPCA model (\ref{eqn1}). 
To the best of our knowledge, no such constraints are employed in the literature for background subtraction in the sparse tensor of the TRPCA framework.

We utilize an ADMM batch-based optimization approach to solve the objective function of the proposed STRPCA model since it has improved convergence and accuracy \cite{boyd2011distributed}.
In several SOTA approaches \cite{lu2019tensor, xue2021multilayer, wang2021tensor, sobral2017matrix}, batch processing is effective, but not for applications that require real-time processing.
As a result, in the current work, we also proposed an online optimization strategy to solve the objective function.
One video frame at a time is processed by our proposed Online STRPCA optimization model, called O-STRPCA, which also concurrently encodes the spatial-temporal regularization.

On six publicly accessible background subtraction benchmark datasets, including Change Detection.Net 2014 (CD14) \cite{wang2014cdnet}, Institute for Infocomm Research (I2R) \cite{li2004statistical}, Background Model Challenges 2012 (BMC12) \cite{vacavant2013benchmark}, Wallflower \cite{toyama1999wallflower}, Stuttgart Artificial Background Subtraction (SABS) \cite{brutzer2011evaluation}, SBM-RGBD \cite{camplani2017benchmarking}, we assessed the performance of our proposed STRPCA and O-STRPCA algorithms.
Our results showed that the proposed algorithms outperformed SOTA techniques for background subtraction.
The following is a summary of our work's significant contributions:

\begin{enumerate}
\item We proposed a novel STRPCA algorithm for enhanced background subtraction.  
Our algorithm applies graph-Laplacian matrices to the sparse tensor $\mathbfcal{F}$ to enforce the spatial and temporal regularizations.
These regularizations support the spatial-temporal coherent moving object structure and handle the structure-sparsity issues.
\item We put forth a novel objective function that concurrently uses both spatial-temporal regularizations and $\mathbfcal{B}$-$\mathbfcal{F}$ tensors. 
Then, batch-based and online-based optimization techniques are used to resolve it. 
Even though our batch solution worked better, real-time applications are better suited to the online solution. 
\item On six publicly accessible datasets, we conducted in-depth qualitative and quantitative analyses and compared the STRPCA performance with that of 15 SOTA approaches. There has also been a thorough review of the outcomes.     
\end{enumerate}

This work is organized as follows. 
The literature review on the RPCA and TRPCA approaches is summarised in Sec. \ref{sec:relatedwork}.
The proposed algorithm is explained in Sec. \ref{sec:method}.
Extensive experiments are presented in Sec. \ref{sec:results} and the conclusion and suggested future directions are presented in Sec. \ref{sec:conclusion} 

\section{Related Work}
\label{sec:relatedwork}
The last two decades have produced a wealth of work on background subtraction, which may be divided into conventional \cite{bouwmans2011recent}, subspace learning \cite{bouwmans2017decomposition, sobral2017matrix}, and deep learning \cite{bouwmans2019deep} approaches.
Below, we provide a concise summary of each background subtraction method category.

\subsubsection{\textbf{Traditional Methods}} 
In this category, pixel-level approaches received a lot of attention for dynamic background subtraction \cite{bouwmans2011recent}.
Stauffer and Grimson proposed a Gaussian Mixture Model (GMM) in which each pixel is modeled using a combination of Gaussian probability density functions\cite{stauffer1999adaptive}.
In the event that the pixel values do not meet the criterion for background distribution, the moving object pixels are then categorized as foregrounds.
Despite the method's optimistic background subtraction performance, it is vulnerable to rapid background fluctuations like the timing of light switches and other factors like the number of Gaussians.
As a result, numerous improved GMM approaches are put forth in the literature.
These approaches include adaptive GMM \cite{zivkovic2004improved}, spanning tree GMM \cite{chen2014spatiotemporal, chen2017spatiotemporal}, bidirectional GMM \cite{shimada2013background}, features-based GMM \cite{shah2014video}, and structured GMM model \cite{shi2018robust}.

SuBSENSE \cite{st2014subsense} and PAWCS \cite{st2016universal} are the most modern pixel-level background subtraction techniques.
Pierre-Luc \textit{et al.} proposed a universal change detection method known as SuBSENSE which is robust against local variations in the background scene.
SuBSENSE, a technique for detecting global changes that is resistant to small fluctuations in the background scene, was suggested by Pierre-Luc \textit{et al.} \cite{st2014subsense}.
SuBSENSE uses spatiotemporal binary and color information to categorize each pixel as either background or foreground.
Based on the local information, the number of parameters is dynamically updated.
For long-term foreground segmentation, Pierre-Luc \textit{et al.} also put out a non-parametric PAWCS technique that learns the static and dynamic background pixels online at a low memory cost \cite{st2016universal}.

\subsubsection{\textbf{Subspace Learning Methods}} 
This group of techniques compels a background model to be linearly correlated and learns a low-dimensional subspace of the input sequence.
Background subtraction techniques based on PCA, RPCA, and TRPCA have seen a lot of success in recent years \cite{bouwmans2017decomposition}.
Using PCA, Oliver \textit{et al.} devised the eigen background subtraction technique \cite{oliver2000bayesian}.
PCA results were optimistic, but the low-dimensional background subspace that was learned was particularly vulnerable to noise or severely distorted outliers in the background scene. 
Wright \textit{et al.} introduced the RPCA for learning both low-rank and sparse subspaces using a convex optimization program to overcome this issue \cite{wright2009robust}.
A nuclear norm minimization was imposed as a convex relaxation in the RPCA model since rank minimization was not continuous and non-convex.
It was used for background subtraction by Cand{\`e}s \textit{et al.} \cite{candes2011robust}.
Classical RPCA methods have shown to be potential solutions for background subtraction however, these methods are computationally not attractive due to batch optimization processing and also can not handle dynamic background scenes.
Therefore, many RPCA variants have been proposed such as DECOLOR \cite{zhou2012moving}, TVRPCA \cite{cao2015total}, LSD \cite{liu2015background}, 2PRPCA \cite{gao2014block}, GOSUS \cite{xu2013gosus}, MSCL \cite{javed2017background}, and DSPSS \cite{ebadi2017foreground} to address the structured-sparsity problem.
Online RPCA variations like ORPCA \cite{javed2015or} and COROLLA \cite{shakeri2016corola} are also published to be able to handle the real-time processing difficulties of RPCA.
Donald \textit{et al.} developed a tensor variation of the robust formulations for RPCA \cite{goldfarb2014robust}.
Recently, Lu \textit{et al.} suggested a TRPCA employing a tensor nuclear norm regularisation to handle multi-dimensional data \cite{lu2019tensor}.
For effective foreground segmentation, Wenrui \textit{et al.} used TRPCA with total variation penalty \cite{hu2016moving}.
A time-lapsed sequence handling model with an invariant tensor sparse decomposition was proposed by Shakeri \textit{et al.} \cite{shakeri2019moving}.
Although TRPCA and its variations show an improvement in SOTA performance over RPCA, their key drawbacks continue to be computational complexity and a lack of sparsity structure.
As a result, online TRPCA versions have also recently been described in the literature \cite{li2022tensor,li2018online}.

A tensor-based subspace learning model for background subtraction is also the foundation of our proposed STRPCA algorithm.
We impose graph-based spatial and temporal continuity to the sparse component and enhance background subtraction performance in comparison to the aforementioned TRPCA approaches.

\textbf{Deep Learning Methods}
Many computer vision applications, including object detection \cite{zhao2019object}, object tracking \cite{javed2022visual}, and background subtraction \cite{bouwmans2019deep}, have been transformed by deep Convolutional Neural Networks (CNNs). 
Convolutional features are trained entirely by a fully supervised CNN model before being used for either classification or regression tasks. 
One of the earliest deep learning techniques for background subtraction was proposed by Braham \textit{et al.} \cite{braham2016deep}. 
Using a fully CNN model, each input frame is separated into blocks, and each block is then categorized as either foreground or background. 
In the same vein, Wang \textit{et al.} suggested an interactive block-based deep neural network for moving object segmentation utilizing AlexNet as a backbone architecture \cite{wang2017interactive}.
Although the performance of these two block-based deep networks was impressive, both models degraded in the presence of unknown classes. 
Tezcan \textit{et al.} proposal for a comprehensive CNN model for pixel-wise background subtraction of unknown sequences is, as a result, \cite{tezcan2020bsuv}. 
Readers who are interested might investigate further deep learning techniques for background subtraction in the survey \cite{bouwmans2019deep}.
One of the major problems of fully supervised CNN models is that they strongly rely on well-annotated, highly diversified, and abundant data, which is not always accessible. 
This is true even though the background subtraction group has made exceptional progress. 
Our proposed algorithm, on the other hand, is based on TRPCA, which completely unsupervisedly divides input video sequences into background-foreground components. 

\begin{figure*}[t!]
\centering
\includegraphics[width=\linewidth, height=4.8in]{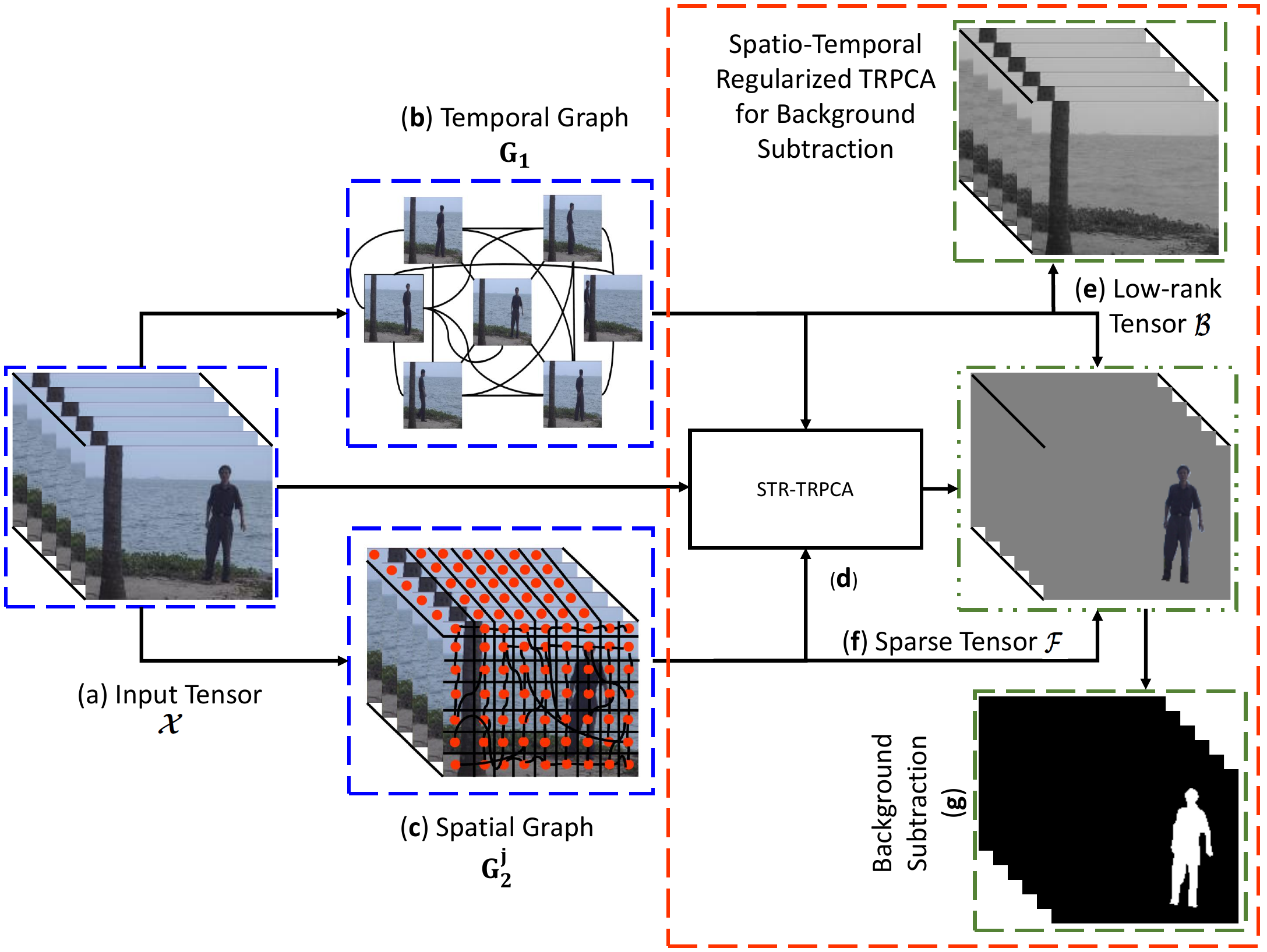}
\caption{Schematic illustration of the proposed STRPCA algorithm for background subtraction. Step (a) shows an input tensor $\mathbfcal{X}$, step (b) shows the construction of temporal graph $\textbf{G}_{1}$, step (c) shows the construction of spatial graph $\textbf{G}_{2}^{j}$, step (d) shows the batch-based STRPCA model optimization where both graphs are incorporated, and steps (e)-(g) show the resulting low-rank tensor $\mathbfcal{B}$, sparse tensor $\mathbfcal{F}$, and the background subtraction results.}
\label{fig_maindiagram}
\end{figure*}

\section{Proposed Methodology}
\label{sec:method}
The system diagram of the proposed Spatial-temporal regularized TRPCA (STRPCA) algorithm for background subtraction is illustrated in Fig. \ref{fig_maindiagram}.
Our proposed algorithm consists of several steps including spatial graph construction, temporal graph construction, objective function formulation, and solution to the proposed model using batch and online optimization techniques.
We initially present mathematical notations and preliminaries before going into detail about each stage of the proposed method.

\subsection{Mathematical Notations and Preliminaries}
In this work, we used the same notations as described in \cite{lu2016tensor, lu2019tensor, goldfarb2014robust, kolda2009tensor}. 
We denote a multi-dimensional tensor comprising of an input video sequence (3-way tensor) by boldface calligraphic letters, e.g., $\mathbfcal{X} \in \mathbb{R}^{w \times h \times n}$, where $w$, $h$, and $n$ represent width, height, and a number of frames.
Matrices are denoted by boldface capital letters e.g., \textbf{X}, vectors are represented by boldface lowercase letters e.g., \textbf{x}, and scalars are represented by lowercase letters e.g., \textit{a}.
The $(i,j,k)$-th entry of a tensor $\mathbfcal{X}$ is denoted by $\mathbfcal{X}(i,j,k)$ or $x_{ijk}$.

\subsubsection{\textbf{Tensor Slices}} The slices of a tensor forms a 2D matrix.
We use the Matlab notations $\mathbfcal{X}(i,:,:)$, $\mathbfcal{X}(:,j,:)$, and $\mathbfcal{X}(:,:,k)$ respectively, to denote the $i$-th horizontal, $j$-th lateral, and $k$-th frontal slice of tensor $\mathbfcal{X}$.
$\textbf{X}^{(k)}$ is also equivalent to $\mathbfcal{X}(:,:,k)$. 

\subsubsection{\textbf{Tensor Unfolding and Folding Operations}}
\label{sec:fold}
The unfolding operation $\textrm{unfold}(\mathbfcal{X})$ turns $\mathbfcal{X}$ into a 2D matrix and the folding operation is its inverse operator. 
For tensor $\mathbfcal{X} \in \mathbb{R}^{w \times h \times n}$, we define its mode-2 unfolding matrix $\mathbfcal{X}_{(2)}$ as: $\textrm{unfold}_{2}(\mathbfcal{X})=[\textbf{X}^{(1)}, \textbf{X}^{(2)},...,\textbf{X}^{(n)}]^{\top}\in \mathbb{R}^{wn \times h}$.
Its folding matrix is defined as $\textrm{fold}(\textrm{unfold}_{2}(\mathbfcal{X}))=\mathbfcal{X}$. 
Similarly, mode-1 and mode-3 unfolding matrices are 
$\mathbfcal{X}_{(1)}=\textrm{unfold}_{1}(\mathbfcal{X})\in\mathbb{R}^{hn \times w}$
and $\mathbfcal{X}_{(3)}=\textrm{unfold}_{3}(\mathbfcal{X})\in\mathbb{R}^{wh \times n}$.

\subsubsection{\textbf{DCT of Tensor}} We denote $\bar{\mathbfcal{X}}$ as a result of the Discrete Fourier Transformation (DFT) of a tensor $\mathbfcal{X}$ along the third dimension and can be computed using a Matlab command \textit{fft} as: $\bar{\mathbfcal{X}}=\textit{fft}(\mathbfcal{X},[~],3)$.  
Similarly, $\bar{\mathbfcal{X}}$ can be transformed back to $\mathbfcal{X}$ using inverse DCT as: $\mathbfcal{X}=\textrm{fft}(\bar{\mathbfcal{X}},[~],3)$. 
\subsubsection{\textbf{Tensor Norms}} We employed three important norms of a tensor including the $\ell_{1}$-norm $\lvert\vert\mathbfcal{X}\vert\rvert_{1}=\sum_{i,j,k}\lvert x_{ijk}\rvert$,
the Frobenius norm $\lvert\vert\mathbfcal{X}\vert\rvert_{F}=\sqrt{(\sum_{i,j,k}\lvert x_{ijk}\rvert^{2})}$, and the nuclear norm of a matrix  $\lvert\lvert\textbf{X}\rvert\rvert_{*}=\sum_{i}\sigma_{i}(\textbf{X})$, where $\sigma_{i}(\textbf{X})$ is the $i$-th singular values of \textbf{X}.
The tensor nuclear norm is estimated using the frontal slices of the input tensor.
Kilmer \textit{et al.} defined the nuclear norm of a tensor $\lvert\lvert\mathbfcal{X}\rvert\rvert_{*}$ as the sum of the matrix nuclear norms of all the frontal slices of $\bar{\mathbfcal{X}}$ as \cite{kilmer2013third}: $\lvert\lvert\mathbfcal{X}\rvert\rvert_{*}=\sum_{i=1}^{n}\lvert\lvert\bar{\mathbfcal{X}}(:,:,i)\rvert\rvert_{*}$.
Lu \textit{et al.} proposed to take an average of all matrix nuclear norms as \cite{lu2016tensor}: $\lvert\lvert\mathbfcal{X}\rvert\rvert_{*}=\frac{1}{n}\sum_{i=1}^{n}\lvert\lvert\bar{\mathbfcal{X}}(:,:,i)\rvert\rvert_{*}$.





\subsubsection{\textbf{Tensor-Tensor Product}}
The tensor-tensor product (t-product) between any two tensors $\mathbfcal{Y}_{1} \in \mathbb{R}^{n_{1} \times n_{2} \times n_{3}}$ and $\mathbfcal{Y}_{2} \in \mathbb{R}^{n_{2} \times c \times n_{3}}$ is defined to be a tensor $\mathbfcal{Z} \in \mathbb{R}^{n_{1} \times c \times n_{3}}$ and is computed as \cite{lu2019tensor}: $\mathbfcal{Z}=\textrm{fold}(\textrm{bcirc}(\mathbfcal{Y}_{1}). \textrm{unfold}(\mathbfcal{Y}_{2}))$, where $\textrm{bcirc}(\mathbfcal{Y}_{1})$ is a block-circulant matrix of size $n_{1}n_{3} \times n_{2}n_{3}$.
\subsubsection{\textbf{Tensor Singular Value Decomposition (T-SVD)}}
\label{sec:svd}
The input tensor $\mathbfcal{X}$ can be factorized as: $\mathbfcal{X}= \mathbfcal{U}*\mathbfcal{S}*\mathbfcal{V}^{*}$, where $\mathbfcal{U}\in \mathbb{R}^{w \times w \times n}$ and $\mathbfcal{V}\in \mathbb{R}^{h \times h \times n}$ are orthogonal tensors, and $\mathbfcal{S}\in \mathbb{R}^{w \times h \times n}$ is an f-diagonal tensor.
$\mathbfcal{V}^{*}$ is the conjugate transpose of tensor $\mathbfcal{V}$.
These factored tensors contain principal components of $\mathbfcal{X}$.
The T-SVD of $\mathbfcal{X}$ can be computed using the below steps as \cite{lu2019tensor}:

\begin{enumerate}
\item  Compute $\bar{\mathbfcal{X}}=\textrm{fft}(\mathbfcal{X},[~],3)$.  \item Compute $[\bar{\textbf{U}}^{(k)},\bar{\textbf{S}}^{(k)},\bar{\textbf{V}}^{(k)}]=\textrm{SVD}(\bar{\textbf{X}}^{(k)})$.
\item Compute complex conjugates of  $\bar{\textbf{U}}^{(k)}$ and $\bar{\textbf{V}}^{(k)}$.
\item Compute $\mathbfcal{U}=\textrm{ifft}(\bar{\mathbfcal{U}},[~],3)$, $\mathbfcal{S}=\textrm{ifft}(\bar{\mathbfcal{S}},[~],3)$, and $\mathbfcal{V}=\textrm{ifft}(\bar{\mathbfcal{V}},[~],3)$.
\end{enumerate}

\noindent The complete collection of symbols and notations used in this work is shown in Table \ref{table1}.

\begin{table*}[t!]
\caption{Description of the notations used in this work.}
\begin{center}
\makebox[\linewidth]{
\scalebox{0.70}{
\begin{tabu}{|c|c|}
\tabucline[2.0pt]{-}
Notation&Description\\\tabucline[2.0pt]{-}
$\mathbfcal{X}$, \textbf{X}, \textbf{x}, and \textit{x}&Tensor, matrix, vector, and scalar.\\\tabucline[0.5pt]{-}
$\mathbfcal{X}(i,:,:), \mathbfcal{X}(:,j,:), \mathbfcal{X}(:,:,k)$& $i$-th horizontal, $j$-th lateral, and $k$-th frontal slices by fixing all but two indices.\\\tabucline[0.5pt]{-}
$\mathbfcal{X}(i,j,k)$ or $x_{i,j,k}$& $(i,j,k)$-th entry of tensor $\mathbfcal{X}$.\\\tabucline[0.5pt]{-}
$\mathbfcal{X}^{(k)}$, $\mathbfcal{X}_{(n)}$, and $\bar{\mathbfcal{X}}$&$k$-th frontal slice,  mode-$n$ matricization, and DCT of $\mathbfcal{X}$.\\\tabucline[0.5pt]{-}
$||\mathbfcal{X}||_{*}$, $||\mathbfcal{X}||_{1}$, and $||\mathbfcal{X}||_{F}$& Nuclear, $l_{1}$, and Frobenius norms of a tensor $\mathbfcal{X}$.\\\tabucline[0.5pt]{-}
$\mathbfcal{B}$ and $\mathbfcal{F}$ &Low-rank and sparse tensors comprising a background and foreground components.\\\tabucline[0.5pt]{-}
$\mathbfcal{H}$ and $\mathbfcal{T}$ &Spatial and temporal moving object tensors.\\\tabucline[0.5pt]{-}
$\textbf{G}_{1}$ and $\textbf{L}_{t}$&Temporal graph and normalized graph-based temporal Laplacian matrix.\\\tabucline[0.5pt]{-}
$\textbf{G}_{s}^{j}$ and $\mathbfcal{L}_{s}$&Spatial graph of $j$-th frontal slice and normalized graph-based spatial Laplacian tensor.\\\tabucline[0.5pt]{-}
$\textbf{x}_{m}^{i}, \textbf{v}_{m}^{i}, \textbf{f}_{m}^{i}, \textbf{h}_{m}^{i}$, and $\textbf{t}_{m}^{i}$& $i$-th sample of the mode-$n$ matricized tensors $\mathbfcal{X}_{(m)}, \mathbfcal{V}_{(m)}, \mathbfcal{F}_{(m)}, \mathbfcal{H}_{(m)}$, and $\mathbfcal{T}_{(m)}$.\\\tabucline[0.5pt]{-}
\end{tabu}
}}
\end{center}
\label{table1}
\end{table*}

\subsection{Mathematical Formulation of STRPCA}
TRPCA (\ref{eqn1}) aims to decompose an input tensor $\mathbfcal{X}$ into the sum of a low-rank component $\mathbfcal{B}$ representing the background model and a sparse component $\mathbfcal{F}$ representing the moving object.
However, the development of spatial-temporal structure within the sparse component $\mathbfcal{F}$ is hindered by the absence of structured-sparsity regularisation in previous TRPCA techniques \cite{sobral2017matrix, hu2016moving, li2022tensor, li2018online, lu2019tensor, lu2016tensor}, resulting in an erroneous background subtraction results.
We propose adding spatial-temporal constraints to the model (\ref{eqn1}) in order to correctly segment the moving object pixels.
The objective function of the proposed STRPCA model is formulated as:

\begin{equation}
\begin{split}
\min_{\mathbfcal{B},\mathbfcal{F}}||\mathbfcal{B}||_{*}+\lambda||\mathbfcal{F}||_{1} +\gamma_{1} \sum_{j=1}^{n} \textrm{Tr}(\mathbfcal{F}^{(j)\top}\mathbfcal{L}^{(j)}_{s}\mathbfcal{F}^{(j)}) \\ + \gamma_{2}  \textrm{Tr}(\mathbfcal{F}_{(3)}\mathbf{L}_{t}\mathbfcal{F}^{\top}_{(3)}), 
\textrm{ such that}~\mathbfcal{X}=\mathbfcal{B}+\mathbfcal{F},
\end{split}
\label{eqn2}
\end{equation}

\noindent where $\textrm{Tr}(.)$ denotes the trace of a matrix.
The third and fourth components are known as spatial and temporal graph-based regularization enforced on $\mathbfcal{F}$.
By computing the pair-wise similarities along the spatial and temporal dimensions, we consider these regularisations to be a search for intact structured moving objects.
The spatial graph-based Laplacian tensor, denoted by the $\mathbfcal{L}_{s}$ is specifically calculated pixel-wise from the spatial graph using the frontal slices of $\mathbfcal{X}$.
By incorporating $\sum_{j=1}^{n} \textrm{Tr}(\mathbfcal{F}^{(j)\top}\mathbfcal{L}^{(j)}_{s}\mathbfcal{F}^{(j)})$ component, we compel the $\mathbfcal{F}$ component to act as the eigenvectors of the $\mathbfcal{L}_{s}$, which guarantees that the moving object's spatial coherent structure is preserved.
$\mathbfcal{L}_{t}$ denotes the temporal graph-based Laplacian matrix computed frame-wise in the temporal domain using the mode-3 unfolded matrix of $\mathbfcal{X}$.
Similarly, by including $\textrm{Tr}(\mathbfcal{F}_{(3)}\mathbf{L}_{t}\mathbfcal{F}^{\top}_{(3)})$ component, we maintain the moving object's temporal coherent structure.
While optimising STRPCA (\ref{eqn3}), the non-negative tradeoff parameters $\gamma_{1}$ and $\gamma_{1}$ determine the degree of moving object sparsity and give relative relevance to each component. 
We introduce spatial $\mathbfcal{H}$ and temporal $\mathbfcal{H}$ moving object tensors as follows to assist (\ref{eqn3}) be more separable:

\begin{equation}
\begin{split}
\min_{\mathbfcal{B},\mathbfcal{F}}||\mathbfcal{B}||_{*}+\lambda||\mathbfcal{F}||_{1} +\gamma_{1} \sum_{j=1}^{n} \textrm{Tr}(\mathbfcal{H}^{(j)\top}\mathbfcal{L}^{(j)}_{s}\mathbfcal{H}^{(j)}) \\ + \gamma_{2}\textrm{Tr}(\mathbfcal{T}_{(3)}\mathbf{L}_{t}\mathbfcal{T}^{\top}_{(3)}), 
\textrm{ such that}~\mathbfcal{X}=\mathbfcal{B}+\mathbfcal{F}, \\
\mathbfcal{H}=\mathbfcal{F},~\textrm{and}~\mathbfcal{T}=\mathbfcal{F}. 
\end{split}
\label{eqn3}
\end{equation}

\noindent To solve (\ref{eqn4}), we first compute the spatial-temporal regularization and then we optimize it using the proposed online and batch-based optimization methods.

\subsection{Spatial-Temporal Graph-based Sparse Regularizations}
\label{sec:constraints}
Model (\ref{eqn1}) failed to take into account the spatio-temporal organization of the sparse component, which left gaps in the segmentation of moving objects. 
We suggest spatial-temporal graph-based regularisation to address this problem, and to do so, we create spatial and temporal graphs.

\subsubsection{Temporal Graph-based Laplacian Regularization}
To construct a temporal graph, we initially transform $\mathbfcal{X}$ into its mode-3 unfolded 2-D matrix using an unfolding operation as: $\mathbfcal{X}_{(3)}=\textrm{unfold}_{3}(\mathbfcal{X})= \left[ \mathbfcal{X}(:,1,:)^{\top}, \mathbfcal{X}(:,2,:)^{\top},\cdot \cdot \cdot,\mathbfcal{X}(:,h,:) \right]^{\top} \in \mathbb{R}^{wh \times n}$ in order to capture the temporal continuity.

An undirected temporal graph, $\textbf{G}_{1}=(\textbf{A}_{1},\textbf{E}_{1})$ is then constructed where the vertices $\textbf{V}_{1}$ define the columns of the unfolded matrix $\mathbfcal{X}_{(3)}$ and $\textbf{A}_{1}$ is the adjacency matrix that encodes the pair-wise similarities between the vertices on the graph. 
The key idea here is that the comparable columns of $\mathbfcal{X}_{(3)}$ connected on the graph $\textbf{G}_{1}$ are probably background components $\mathbfcal{B}$, whereas columns that are disconnected or separated from one another on the $\textbf{G}_{1}$ are distinct resulting in a segmentation of moving objects that is temporally consistent.  \cite{yin2015laplacian}.
A graph can be constructed using a variety of methods. 
Due to its ease of use and simplicity, we generate graph using the $k$-Nearest Neighbors (kNN) approach.
Larger graphs are built using the FLANN package for large-scale datasets \cite{muja2014scalable}.
The first step is to find the nearest neighbors for all the vertices using the Euclidean distance, where each vertex is connected to its $k$ closest neighbors.
The adjacency matrix $\textbf{A}_{1}$ for $\textbf{G}_{1}$ holding the pair-wise vertex similarity $a_{1}(p,q)$ between two vertices $p$ and $q$ is then estimated as:

\begin{equation}
a_{1}(p,q)= \exp \Big(\frac{\lvert\lvert \mathbfcal{X}_{(3)}(:,p)-\mathbfcal{X}_{(3)}(:,q)\rvert\rvert_{2}^{2}}{2\sigma_{1}^{2}}\Big).
\label{eqn4}
\end{equation}

\noindent where $\sigma_{1}$ is the temporally smoothing parameter that is estimated using the average distance among the vertices.
Two vertices on the graph $\textbf{G}_{1}$ are connected together if there is an edge between them otherwise $a_{1}(p,q)=0$.
The temporal graph-based regularization on $\mathbfcal{F}_{(3)}$ is given by:

\begin{equation}
\begin{split}
&{}\min_{\mathbfcal{F}_{(3)}} \frac{1}{2} \sum_{p,q=1}^{n} \lvert\lvert \mathbfcal{F}_{(3)}(:,p) - \mathbfcal{F}_{(3)}(:,q)\rvert\rvert_{F}^{2} a_{1}(p,q)\\
&{}=\min_{\mathbfcal{F}_{(3)}} \sum_{p=1}^{n} \mathbfcal{F}_{(3)}(:,p)^{\top}\mathbfcal{F}_{(3)}(:,p)d(p,p)\\
&{}-\sum_{p,q=1}^{n} \mathbfcal{F}_{(3)}(:,p)^{\top}\mathbfcal{F}_{(3)}(:,q)a_{1}(p,q)\\
&{}=\min_{\mathbfcal{F}_{(3)}}\textrm{Tr}(\mathbfcal{F}_{(3)}^{\top}\textbf{D}\mathbfcal{F}_{(3)}) - \textrm{Tr}(\mathbfcal{F}_{(3)}^{\top}\textbf{A}_{1}\mathbfcal{F}_{(3)}) \\
&{}=\min_{\mathbfcal{F}_{(3)}}\textrm{Tr}(\mathbfcal{F}_{(3)}^{\top}\textbf{L}_{t}\mathbfcal{F}_{(3)})
\label{eqn5}
\end{split}
\end{equation} 

\noindent where \textbf{D} is a degree matrix whose diagonal entry is defined as $d(p,p)=\sum_{q}a_{1}(p,q)$ and $\textbf{L}_{t}$ is the normalized graph-based temporal Laplacian matrix which is estimated as:

\begin{equation}
\textbf{L}_{t}= \textbf{I}-\textbf{D}^{-\frac{1}{2}}\textbf{A}_{1}\textbf{D}^{-\frac{1}{2}}.
\label{eqn6}
\end{equation}

\noindent where \textbf{I} is an identity matrix.
The above formulations (\ref{eqn6})-(\ref{eqn7}) encode the temporal structure of the moving object within the sparse tensor $\mathbfcal{F}$. It is interpreted as constraining the $\mathbfcal{F}_{(3)}$ component to be orthogonal to the eigenvectors of normalized graph-based temporal Laplacian matrix.

\subsubsection{Spatial Graph-based Laplacian Regularization}
We construct a pixel-wise spatial graph using the frontal slices of $\mathbfcal{X}$, to maintain the spatial structure of the moving object.
The spatial graph enforces smoothness at pixel-level of each frontal slice in $\mathbfcal{F}$ in contrast to temporal graph regularisation, leading to a spatially coherent segmentation of moving objects.

We initially divide the $j$-th frontal slice $\mathbfcal{X}^{(j)}$ of $\mathbfcal{X}$ into non-overlapping $a \times a$ patches before constructing a spatial graph $\textbf{G}^{j}_{2}=(\textbf{V}^{j}_{2},\textbf{A}^{j}_{2})$ for the $j$-th slice. 
As seen in Fig. \ref{fig_maindiagram} (c), we take a non-overlapping patch for each $i$-th pixel in the $j$-th frontal slice, keeping the $i$-th pixel in the middle.
With $u$ being the total number of pixels in the $j$-th frontal slice, for example, $u=wh$, the spatial data matrix $\textbf{P}^{j}_{s}$ of size $a^{2} \times u$ for the $j$-th frontal slice is produced in this fashion.
All local patches are represented by the vertices $\textbf{V}^{j}_{2}$ in $\textbf{G}^{j}_{2}$, and the $j$-th frontal slice adjacency matrix, which contains all pair-wise similarities between the local patches of each frontal slice, is represented by $\textbf{A}^{j}_{2}$. 

Here, $\textbf{G}^{j}_{2}$ refines the spatial organization of the moving objects and completes the data gathered by $\textbf{G}_{1}$.
In particular, if two local patches are related to one another in $\textbf{G}^{j}_{2}$, their structures are presumably comparable and they are most likely background component pixels. 
We create $\textbf{G}^{j}_{2}$ by measuring the Euclidean distance between the nearby patches and identifying their nearest neighbors.  
The following is an estimate for the adjacency matrix $\textbf{A}^{j}_{2}$ of $\textbf{G}^{j}_{2}$:

\begin{equation}
a_{2}^{j}(p,q)= \begin{cases} \exp \Big(\frac{\lvert\lvert \textbf{P}^{j}_{s}(:,p)-\textbf{P}^{j}_{s}(:,q)\rvert\rvert_{2}^{2}}{2\sigma_{2}^{2}}\Big),~\textrm{if both are connected} \\
0, \textrm{otherwise},
\end{cases}
\label{eqn7}
\end{equation}

\noindent where $\sigma_{2}$  is a smoothing factor for $\textbf{G}^{j}_{2}$ that is calculated using the average distance between vertices.
In case, two local patches $\textbf{P}^{j}_{s}(:, p)$ and $\textbf{P}^{j}_{s}(:, q)$ are connected to each other, then the Euclidean distance is greater than zeros otherwise $a_{2}^{j}(i,j)=0$.
Similar to Eq. (\ref{eqn7}), we compute $\mathbfcal{L}_{s}$ and encode it in $\mathbfcal{F}$.
\vspace{-5mm}
\subsection{Batch Optimization of STRPCA Model}
We employ an ADMM method to solve the model (\ref{eqn4}) in a batch fashion. 
ADMM decomposes the problem into sub-problems and solves each one individually \cite{boyd2011distributed}. 
By eliminating the linear equality constraints, the Lagrangian formulation for (\ref{eqn4}) may be derived as follows:

\begin{equation}
\begin{split}
\mathcal{L}(\mathbfcal{B}, \mathbfcal{F}, \mathbfcal{H}, \mathbfcal{T}, \mathbfcal{Y}_{1}, \mathbfcal{Y}_{2}, \mathbfcal{Y}_{3}, \mu)=
||\mathbfcal{B}||_{*}+\lambda||\mathbfcal{F}||_{1} + \\
\gamma_{1} \sum_{j=1}^{n} \textrm{Tr}(\mathbfcal{H}^{(j)\top}\mathbfcal{L}^{(j)}_{s}\mathbfcal{H}^{(j)}) + \gamma_{2}\textrm{Tr}(\mathbfcal{T}_{(3)}\mathbf{L}_{t}\mathbfcal{T}^{\top}_{(3)}) + \\
\langle  \mathbfcal{Y}_{1},\mathbfcal{X}- \mathbfcal{B}- \mathbfcal{F}\rangle +\frac{\mu}{2}\lvert\lvert\mathbfcal{X}- \mathbfcal{B}- \mathbfcal{F}\rvert\rvert_{F}^{2} + \\
\langle  \mathbfcal{Y}_{2},\mathbfcal{H}-\mathbfcal{F}\rangle +\frac{\mu}{2}\lvert\lvert\mathbfcal{H}-\mathbfcal{F}\rvert\rvert_{F}^{2} + \\
\langle  \mathbfcal{Y}_{3},\mathbfcal{T}-\mathbfcal{F}\rangle +\frac{\mu}{2}\lvert\lvert\mathbfcal{T}-\mathbfcal{F}\rvert\rvert_{F}^{2}, 
\end{split}
\label{eqn8}
\end{equation}

\noindent where  $\mathbfcal{Y}_{1}, \mathbfcal{Y}_{2}$, and $\mathbfcal{Y}_{3}$ are tensors of Lagrangian multipliers and $\mu>0$ is the penalty operator. 
While solving for (\ref{eqn8}), each iteration updates each of these terms. 
The above formulation can also be written as:

\begin{equation}
\begin{split}
\mathcal{L}(\mathbfcal{B}, \mathbfcal{F}, \mathbfcal{H}, \mathbfcal{T}, \mathbfcal{Y}_{1}, \mathbfcal{Y}_{2}, \mathbfcal{Y}_{3}, \mu)=
||\mathbfcal{B}||_{*}+\lambda||\mathbfcal{F}||_{1} + \\
\gamma_{1} \sum_{j=1}^{n} \textrm{Tr}(\mathbfcal{H}^{(j)\top}\mathbfcal{L}^{(j)}_{s}\mathbfcal{H}^{(j)}) + \gamma_{2}\textrm{Tr}(\mathbfcal{T}_{(3)}\mathbf{L}_{t}\mathbfcal{T}^{\top}_{(3)}) + \\
\frac{\mu}{2}\Big\lvert\Big\lvert\mathbfcal{B}+ \mathbfcal{F}- \mathbfcal{X}-\frac{\mathbfcal{Y}_{1}}{\mu}\Big\rvert\Big\rvert_{F}^{2} +
\frac{\mu}{2}\Big\lvert\Big\lvert\mathbfcal{F}-\mathbfcal{H}- \frac{\mathbfcal{Y}_{2}}{\mu}\Big\rvert\Big\rvert_{F}^{2}  \\
+ \frac{\mu}{2}\Big\lvert\Big\lvert\mathbfcal{F}-\mathbfcal{T}- \frac{\mathbfcal{Y}_{3}}{\mu}\Big\rvert\Big\rvert_{F}^{2}, 
\end{split}
\label{eqn9}
\end{equation}

\noindent The solution for each tensor $\mathbfcal{B}, \mathbfcal{F}, \mathbfcal{H}, \mathbfcal{T}, \mathbfcal{Y}_{1}, \mathbfcal{Y}_{2}$, and $\mathbfcal{Y}_{3}$ is then formulated by fixing one tensor and solving another.

\subsubsection{\textbf{Solution for $\mathbfcal{B}$}} Fixing other tensors, a sub-problem $\mathbfcal{B}$ can be updated as follows:

\begin{equation}
\begin{split}
&{}\mathbfcal{B}=\argmin_{\mathbfcal{B}} \mathcal{L}(\mathbfcal{B})=
\argmin_{\mathbfcal{B}} ||\mathbfcal{B}||_{*}+ \frac{\mu}{2}\Big\lvert\Big\lvert\mathbfcal{B}+ \mathbfcal{F}- \mathbfcal{X}-\frac{\mathbfcal{Y}_{1}}{\mu}\Big\rvert\Big\rvert_{F}^{2} \\
&{}=\argmin_{\mathbfcal{B}} \tau ||\mathbfcal{B}||_{*}+ \frac{1}{2}\Big\lvert\Big\lvert\mathbfcal{B}- \mathbfcal{Z}\Big\rvert\Big\rvert_{F}^{2},
\end{split}
\label{eqn10}
\end{equation}

\noindent where $\tau=1/\mu$ and $\mathbfcal{Z}=\mathbfcal{X}-\mathbfcal{F}+\mathbfcal{Y}_{1}/\mu$.
The closed-form solution to sub-problem $\mathbfcal{B}$ is obtained using the tensor Singular Value Thresholding (t-SVT) operation \cite{lu2019tensor} of $\mathbfcal{Z}$ as:

\begin{equation}
\mathbfcal{Z}= \mathbfcal{U} * \mathbfcal{S}_{\tau} * \mathbfcal{V}^{*}, \\
\mathbfcal{S}_{\tau}= \textrm{ifft}((\bar{\mathbfcal{S}}-\tau)_{+}, [~ ],3),
\label{eqn11}
\end{equation}

\noindent where $(\bar{\mathbfcal{S}}-\tau)_{+}$ represents the positive part of $(\bar{\mathbfcal{S}}-\tau)_{+}$ and $\mathbfcal{U}$, $\mathbfcal{S}$, and $\mathbfcal{V}^{*}$ is the T-SVD as defined in Sec. ( \ref{sec:svd}).
For $k+1$ iteration, $\mathbfcal{B}^{k+1}$ is estimated as:

\begin{equation}
\begin{split}
\mathbfcal{B}^{k+1} = \argmin_{\mathbfcal{B}} \tau ||\mathbfcal{B}^{k}||_{*}+ \frac{1}{2}\Big\lvert\Big\lvert\mathbfcal{B}^k- \mathbfcal{Z}^{k}\Big\rvert\Big\rvert_{F}^{2}.
\end{split}
\label{eqn12}
\end{equation}

\subsubsection{\textbf{Solution for $\mathbfcal{T}$}} Fixing other tensor variables, a solution to the sub-problem $\mathbfcal{T}$ is then formulated as:

\begin{equation}
\begin{split}
\mathbfcal{T}=\argmin_{\mathbfcal{T}}\gamma_{2}\textrm{Tr}(\mathbfcal{T}_{(3)}\mathbf{L}_{t}\mathbfcal{T}^{\top}_{(3)}) 
+\frac{\mu}{2}\Big\lvert\Big\lvert\mathbfcal{F}-\mathbfcal{T}- \frac{\mathbfcal{Y}_{3}}{\mu}\Big\rvert\Big\rvert_{F}^{2}.
\end{split}
\label{eqn13}
\end{equation}

\noindent since the computation of $\mathbfcal{T}_{(3)}$ is based on mode-3 unfolded 2-D matrix, therefore we convert all other tensors in  (\ref{eqn13}) as:

\begin{equation}
\begin{split}
\mathbfcal{T}=\argmin_{\mathbfcal{T}}\gamma_{2}\textrm{Tr}(\mathbfcal{T}_{(3)}\mathbf{L}_{t}\mathbfcal{T}^{\top}_{(3)}) 
+\frac{\mu}{2}\Big\lvert\Big\lvert\mathbfcal{T}_{(3)}+ \frac{\mathbfcal{Y}_{3(3)}}{\mu}-\mathbfcal{F}_{(3)}\Big\rvert\Big\rvert_{F}^{2}.
\end{split}
\label{eqn14}
\end{equation}

\noindent by taking the derivative with respect to $\mathbfcal{T}_{(3)}$ and setting its gradient to zero in Eq. (\ref{eqn14}) becomes

\begin{equation}
\gamma_{2}\mathbfcal{T}_{(3)}\mathbf{L}_{t}+\gamma_{2}\mathbfcal{T}_{(3)}\mathbf{L}_{t}^{\top}+\mu \mathbfcal{T}_{(3)}+\mu \mathbfcal{Y}_{3(3)}-\mu \mathbfcal{F}_{(3)}=0.
    \label{eqn15}
\end{equation}

\noindent Finally, the solution for $\mathbfcal{T}_{(3)}$ is given by:

\begin{equation}
\mathbfcal{T}_{(3)}=\frac{\mu \mathbfcal{F}_{(3)}-\mathbfcal{Y}_{3(3)}}{\gamma_{2}\mathbf{L}_{t}+\gamma_{2}\mathbf{L}_{t}^{\top}+\mu\textbf{I}}.
    \label{eqn16}
\end{equation}

\noindent For $k+1$ iteration, $\mathbfcal{T}_{(3)}^{k+1}$ is estimated as:

\begin{equation}
\mathbfcal{T}_{(3)}^{k+1}=\frac{\mu \mathbfcal{F}^{k+1}_{(3)}-\mathbfcal{Y}^{k}_{3(3)}}{\gamma_{2}\mathbf{L}_{t}+\gamma_{2}\mathbf{L}_{t}^{\top}+\mu^{k}\textbf{I}}.
    \label{eqn17}
\end{equation}

\noindent $\mathbfcal{T}_{(3)}^{k+1}$ is converted back to tensor using $\mathbfcal{T}=\textrm{fold}(\mathbfcal{T}_{(3)}^{k+1})$ as defined in Sec. (\ref{sec:fold}).

\subsubsection{\textbf{Solution for $\mathbfcal{H}$}} Fixing other tensors that do not depends on $\mathbfcal{H}$, a solution is then formulated as follows:

\begin{equation}
\begin{split}
\mathbfcal{H}=\argmin_{\mathbfcal{H}}\gamma_{1} \sum_{j=1}^{n} \textrm{Tr}(\mathbfcal{H}^{(j)\top}\mathbfcal{L}^{(j)}_{s}\mathbfcal{H}^{(j)})  
+ \frac{\mu}{2}\Big\lvert\Big\lvert\mathbfcal{F}-\mathbfcal{H}- \frac{\mathbfcal{Y}_{2}}{\mu}\Big\rvert\Big\rvert_{F}^{2}. 
\end{split}
\label{eqn18}
\end{equation}

\noindent Since the computation of $\mathbfcal{H}$ is based on each frontal slice, we also compute other tensors based on frontal slices  in (\ref{eqn18}) as:

\begin{equation}
\begin{split}
\mathbfcal{H}=\argmin_{\mathbfcal{H}}\gamma_{1} \sum_{j=1}^{n} \textrm{Tr}(\mathbfcal{H}^{(j)\top}\mathbfcal{L}^{(j)}_{s}\mathbfcal{H}^{(j)}) \\
+ \frac{\mu}{2}\Big\lvert\Big\lvert \sum_{j=1}^{n} \Big(\mathbfcal{H}^{(j)}+ \frac{\mathbfcal{Y}^{(j)}_{2}}{\mu}-\mathbfcal{F}^{(j)}\Big)\Big\rvert\Big\rvert_{F}^{2}. 
\end{split}
\label{eqn19}
\end{equation}

\noindent by taking the derivative and setting it to zero, (\ref{eqn19}) becomes

\begin{equation}
\gamma_{1} \sum_{j=1}^{n}\Big(\mathbfcal{H}^{(j)}\mathbfcal{L}^{(j)}_{s}+\mathbfcal{H}^{(j)}\mathbfcal{L}^{\top(j)}_{s}\Big)+\mu  \sum_{j=1}^{n}\Big(\mathbfcal{H}^{(j)}+ \mathbfcal{Y}_{2}^{(j)}-\mathbfcal{F}^{(j)}\Big)=0.
    \label{eqn20}
\end{equation}

\begin{equation}
\mathbfcal{H}^{(j)}=\frac{\mu \sum_{j=1}^{n}\Big(\mathbfcal{F}^{(j)}-\mathbfcal{Y}_{2}^{(j)}/\mu\Big)}{2\gamma_{1}\sum_{j=1}^{n}\mathbfcal{L}_{s}^{(j)}+\mu\textbf{I}}.
    \label{eqn21}
\end{equation}

\noindent For $k+1$ iteration, $\mathbfcal{H}^{(j)k+1}$ is estimated as:

\begin{equation}
\mathbfcal{H}^{(j)k+1}=\frac{\mu \sum_{j=1}^{n}\Big(\mathbfcal{F}^{(j)k+1}-\mathbfcal{Y}_{2}^{(j)k}/\mu\Big)}{2\gamma_{1}\sum_{j=1}^{n}\mathbfcal{L}_{s}^{(j)}+\mu^{k}\textbf{I}}.
    \label{eqn22}
\end{equation}

\subsubsection{\textbf{Solution for $\mathbfcal{F}$}} 
Fixing other tensors, a solution to a sub-problem $\mathbfcal{F}$ is then formulated as follows:


\begin{equation}
\begin{split}
\mathbfcal{F}=\argmin_{\mathbfcal{F}}\lambda||\mathbfcal{F}||_{1}+\frac{1}{2}\lvert\lvert \mathbfcal{F} - \mathbfcal{Z}\rvert\rvert_{F}^{2},
\end{split}
\label{eqn23}
\end{equation}

\noindent where $\mathbfcal{Z}=\mathbfcal{X}-\mathbfcal{B}+\mathbfcal{H}+\mathbfcal{T}+ \Big(\mathbfcal{Y}_{1}+\mathbfcal{Y}_{2}+\mathbfcal{Y}_{3}\Big)/\mu$.
For $k+1$ iteration, $\mathbfcal{F}^{k+1}$ can be updated as:

\begin{equation}
\begin{split}
\mathbfcal{F}^{k+1}=\argmin_{\mathbfcal{F}}\lambda||\mathbfcal{F}^{k}||_{1}+\frac{1}{2}\lvert\lvert \mathbfcal{F}^{k} - \mathbfcal{Z}^{k}\rvert\rvert_{F}^{2},
\end{split}
\label{eqn24}
\end{equation}

\noindent Then, the closed-form solution of $\mathbfcal{F}^{k+1}$ can be obtained using soft-thresholding operator as \cite{lu2019tensor}: $\mathbfcal{F}^{k+1}= \textrm{T}_{\frac{\lambda}{\mu^{k}}}\big(\mathbfcal{Z}^{k}\big)$, where $\textrm{T}_{\frac{\lambda}{\mu^{k}}}(\mathbfcal{Z}^{k})$ is the sof-thresholding operation on tensor $\mathbfcal{Z}^{k}$. It is defined as: $\textrm{T}_{\frac{\lambda}{\mu^{k}}}(\mathbfcal{Z}^{k})=\textrm{sign}(   (\mathbfcal{Z}^{k})_{i,j,k}) \bullet \max(|\mathbfcal{Z}^{k}_{i,j,k}|-\lambda/\mu^{k},0)$.

\subsubsection{\textbf{Updating the Lagrangian Multipliers and $\mu$}} Tensors $\mathbfcal{Y}_{1}$,$\mathbfcal{Y}_{2}$,$\mathbfcal{Y}_{3}$, and parameter $\mu$ are updated as:

\begin{equation}
\begin{split}
&{}\mathbfcal{Y}^{k+1}_{1}= \mathbfcal{Y}^{k}_{1}+\mu^{k} (\mathbfcal{X}-\mathbfcal{B}^{k+1}-\mathbfcal{F}^{k+1}); \\
&{}\mathbfcal{Y}^{k+1}_{2}= \mathbfcal{Y}^{k}_{2}+\mu^{k} (\mathbfcal{H}^{k+1}-\mathbfcal{F}^{k+1}); \\
&{}\mathbfcal{Y}^{k+1}_{3}= \mathbfcal{Y}^{k}_{3}+\mu^{k} (\mathbfcal{T}^{k+1}-\mathbfcal{F}^{k+1}); \mu^{k+1}=\min (\rho\mu^{k}, \mu_{max}),
\end{split}
\label{eqn25}
\end{equation}

\noindent where $\rho$  is a non-negative parameter. 

\subsubsection{\textbf{Convergence Conditions}}
Following convergence criteria is defined according to the KKT condition as \cite{boyd2011distributed,  lu2019tensor}: $\rho_{1} \le \zeta~\&~\rho_{2} \le \zeta~\&~\rho_{3} \le \zeta~\&~ \rho_{4}  \le \zeta~\& \rho_{5} \le \zeta$, where  $\rho_{1} \leftarrow{\lvert\lvert\mathbfcal{X}-\mathbfcal{B}^{k+1}-\mathbfcal{F}^{k+1}\rvert\rvert_{F}^{2}}$, $\rho_{2} \leftarrow{\lvert\lvert \mathbfcal{B}^{k}-\mathbfcal{B}^{k+1}\rvert\rvert_{F}^{2}}$,  $\rho_{3} \leftarrow{\lvert\lvert \mathbfcal{F}^{k}-\mathbfcal{F}^{k+1}\rvert\rvert_{F}^{2}}$, $\rho_{4} \leftarrow{\lvert\lvert \mathbfcal{F}^{k}-\mathbfcal{H}^{k}\rvert\rvert_{F}^{2}}$, and $\rho_{5} \leftarrow{\lvert\lvert \mathbfcal{F}^{k}-\mathbfcal{T}^{k}\rvert\rvert_{F}^{2}}$. $\zeta$ is the tolerance factor that controls the convergence criteria.
Algorithm \ref{algo1} summarizes the main steps.

\begin{algorithm}[t!]
	\SetAlgoLined
	\KwIn{$\mathbfcal{X}$, $\gamma_{1} >0$, $\gamma_{2}>0$, $\lambda$, $\mathbfcal{L}_{s}$ \& $\textbf{L}_{t}$ using (\ref{eqn5})-(\ref{eqn7}).}
	\textbf{Initialization:} $\mu^0=0.01, \mu_{max}=10, \rho=1.2,$\\
$\zeta=0.001, \{ \mathbfcal{B}^{0}, \mathbfcal{F}^{0}, \mathbfcal{T}^{0}, \mathbfcal{H}^{0}, \mathbfcal{Y}^{0}_{1}, \mathbfcal{Y}^{0}_{2}, \mathbfcal{Y}^{0}_{3}\}=0$.\\ 
	\While{$\textrm{not converged}~(k=0,1,..)$}{
	1. Update $\mathbfcal{B}^{k+1}$ using (\ref{eqn10})-(\ref{eqn12}).\\  
	2. Update $\mathbfcal{T}_{3}^{k+1}$ using (\ref{eqn17}) and $\mathbfcal{T}=\textrm{fold}(\mathbfcal{T}_{(3)}^{k+1})$. \\
        3. Update $\mathbfcal{H}^{(j)k+1}$ using (\ref{eqn22}). \\
	4. Update $\mathbfcal{F}^{k+1}$ using (\ref{eqn24}).\\
        5. Update $\mu^{k+1}$ and $\{\mathbfcal{Y}^{k+1}_{j}\}_{j=1}^{3}$ using (\ref{eqn25}). \\
        6. Check convergence. \\  
			}
	\KwOut{$\mathbfcal{B}^{k+1}, \mathbfcal{F}^{k+1}$}
\caption{Pseudo-code of STRPCA model.} 
	\label{algo1}
\end{algorithm}

\subsection{Online Optimization of STRPCA (O-STRPCA) Model}

Although batch optimization is effective in terms of efficiency, it adds computational complexity.
Model (\ref{eqn3}), which requires that all video frames be stored in memory, cannot always be accomplished, especially for real-time processing.
In order to fill this gap, Feng \textit{et al.} suggested an online stochastic optimization approach that processes one frame per time instance \cite{feng2013online}. 
The computational complexity issues are therefore resolved.
However, it lacks structural constraints and is only used for matrix-based RPCA problems.
By using an online optimization technique where one sequence from the input tensor is processed each time occurrence, we are able to solve the model (\ref{eqn3}) at hand.
We proposed O-STRPCA model that includes tensor-based spatial and temporal graph-based Laplacian regularizations in contrast to \cite{feng2013online}. 

To achieve this, we first unfold the model (\ref{eqn3}) before utilizing an online optimization to solve each unfolded matrix.
Model (\ref{eqn3}) unfolded may be represented as follows:

\begin{equation}
\begin{split}
&{}\min_{\begin{subxarray}\mathbfcal{B}_{(m)},\mathbfcal{F}_{(m)}&\\m=1,2,\cdot\cdot\cdot,M\end{subxarray}}\sum_{m=1}^{M} \Big(||\mathbfcal{B}_{(m)}||_{*}+\lambda||\mathbfcal{F}_{(m)}||_{1}+ \gamma_{1}\textrm{Tr}(\mathbfcal{H}^{\top}_{(m)}\mathbfcal{L}_{s(m)} \\
&{}\mathbfcal{H}_{(m)})+ \gamma_{2}\textrm{Tr}(\mathbfcal{T}_{(m)}\mathbf{L}_{t}\mathbfcal{T}^{\top}_{(m)})\Big), \textrm{ such that}~\mathbfcal{X}_{(m)}=\mathbfcal{B}_{(m)}+\mathbfcal{F}_{(m)},  \\
&{}\mathbfcal{H}_{(m)}=\mathbfcal{F}_{(m)},~\mathbfcal{T}_{(m)}=\mathbfcal{F}_{(m)},~m=1,2,\cdot\cdot\cdot,M,~\textrm{and} \\
&{}\{ \mathbfcal{X}, \mathbfcal{B}, \mathbfcal{F}\}=\{\textrm{fold}(\mathbfcal{X}_{(m)}), \textrm{fold}(\mathbfcal{B}_{(m)}), \textrm{fold}(\mathbfcal{F}_{(m)})\},
\end{split}
\label{eqn26}
\end{equation}

\noindent where $M$ is the total number of modes in a tensor. 
To solve (\ref{eqn26}), we first convert it to an unconstrained problem as:

\begin{equation}
\begin{split}
&{}\min_{\begin{subxarray}\mathbfcal{B}_{(m)},\mathbfcal{F}_{(m)}&\\m=1,2,\cdot\cdot\cdot,M\end{subxarray}}\sum_{m=1}^{M} \Big(||\mathbfcal{B}_{(m)}||_{*}+\lambda||\mathbfcal{F}_{(m)}||_{1}+ \gamma_{1}\textrm{Tr}(\mathbfcal{H}^{\top}_{(m)}\mathbfcal{L}_{s(m)} \\
&{}\mathbfcal{H}_{(m)})+ \gamma_{2}\textrm{Tr}(\mathbfcal{T}_{(m)}\mathbf{L}_{t}\mathbfcal{T}^{\top}_{(m)})+ ||\mathbfcal{X}_{(m)}-\mathbfcal{B}_{(m)}-\mathbfcal{F}_{(m)}||_{F}^{2}  + \\
&{}||\mathbfcal{H}_{(m)}-\mathbfcal{F}_{(m)}||_{F}^{2}  + ||\mathbfcal{T}_{(m)}-\mathbfcal{F}_{(m)}||_{F}^{2} \Big).
\end{split}
\label{eqn27}
\end{equation}

\noindent The nuclear norm minimization problem, where SVD is estimated in each iteration and strongly links all principal components, is a significant obstacle to solving the model (\ref{eqn27}) in an online manner.
We use an approximation of the nuclear norm calculated using the matrix factorization problem to fill up this gap \cite{feng2013online}.
The equivalent nuclear norm is the sum of the basis $\mathbfcal{U}$ and its coefficients $\mathbfcal{V}$ and it can be expressed as:

\begin{equation}
\begin{split}
||\mathbfcal{B}_{(m)}||_{*}=\min_{\mathbfcal{U}_{(m)}\in \mathbb{R}^{p \times r}, \mathbfcal{V}_{(m)}\in \mathbb{R}^{q \times r} }\frac{1}{2}\{ ||\mathbfcal{U}_{(m)}||_{F}^{2}+||\mathbfcal{V}_{(m)}||_{F}^{2}\},  \\
 \textrm{ such that}~\mathbfcal{B}_{(m)}=\mathbfcal{U}_{(m)}\mathbfcal{V}_{(m)},
\end{split}
\label{eqn28}
\end{equation}

\noindent where $p$ denotes the dimension of each sample in $\mathbfcal{U}$, $r$ is a rank, and $q$ is the number of samples in $\mathbfcal{V}_{(m)}$.
It should be noted that the dimensions of each unfolded matrices $\mathbfcal{U}_{(m)}$ and $\mathbfcal{V}_{(m)}$ vary according to the size of $\mathbfcal{B}_{(m)}$.
By substituting Eq. (\ref{eqn28}) into Eq. (\ref{eqn27}), we get


\begin{equation}
\begin{split}
&{}\min_{\begin{subxarray}\mathbfcal{U}_{(m)}, \mathbfcal{V}_{(m)}\mathbfcal{F}_{(m)}&\\m=1,2,\cdot\cdot\cdot,M\end{subxarray}}\sum_{m=1}^{M}\Big(||\mathbfcal{X}_{(m)}-\mathbfcal{U}_{(m)}\mathbfcal{V}_{(m)}-\mathbfcal{F}_{(m)}||_{F}^{2}+||\mathbfcal{H}_{(m)}-\\
&{}\mathbfcal{F}_{(m)}||_{F}^{2} +||\mathbfcal{T}_{(m)}-\mathbfcal{F}_{(m)}||_{F}^{2} +\frac{1}{2}\{ ||\mathbfcal{U}_{(m)}||_{F}^{2}+||\mathbfcal{V}_{(m)}||_{F}^{2}\} + \\
&{}\lambda||\mathbfcal{F}_{(m)}||_{1} + \gamma_{1}\textrm{Tr}(\mathbfcal{H}^{\top}_{(m)}\mathbfcal{L}_{s(m)}\mathbfcal{H}_{(m)}) + \gamma_{2}\textrm{Tr}(\mathbfcal{T}_{(m)}\mathbf{L}_{t}\mathbfcal{T}^{\top}_{(m)})\Big).
 \end{split}
\label{eqn29}
\end{equation}

\noindent The aforementioned composition continues to function batch-wise.
We introduce vectorized samples of each tensor and provide an online optimization framework for processing each frame per-time instance in order to convert it to the online formulation as:

\begin{equation}
\begin{split}
&{}\min_{\mathbfcal{U}_{(m)}, \textbf{v}_{m}^{i}\textbf{f}_{m}^{i}}\sum_{i=1}^{N}\Bigg( \sum_{m=1}^{M}||\textbf{x}_{m}^{i}-\mathbfcal{U}_{(m)}\textbf{v}_{m}^{i}-\textbf{f}_{m}^{i}||_{F}^{2} + ||\textbf{h}_{m}^{i}-\textbf{f}_{m}^{i}||_{F}^{2}\\
&{}+ \lambda||\textbf{f}_{m}^{i}||_{1} +||\textbf{t}_{m}^{i}-\textbf{f}_{m}^{i}||_{F}^{2} +\frac{1}{2}||\textbf{v}_{m}^{i}||_{F}^{2} +  \gamma_{1}\textrm{Tr}(\textbf{h}^{i\top}_{m}\mathbfcal{L}_{s(m)}^{i}\textbf{h}_{m}^{i}) + \\
&{}\gamma_{2}\textrm{Tr}(\textbf{t}_{m}^{i}\mathbf{L}_{t}^{i}\textbf{t}^{i \top}_{m})\Bigg) +  \frac{1}{2}\sum_{m=1}^{M}||\mathbfcal{U}_{(m)}||_{F}^{2},
\end{split}
\label{eqn30}
\end{equation}

\noindent where $\textbf{x}_{m}^{i}, \textbf{v}_{m}^{i}, \textbf{f}_{m}^{i}, \textbf{h}_{m}^{i}$, and $\textbf{t}_{m}^{i}$ represents the columns in matrices $\mathbfcal{X}_{(m)}, \mathbfcal{V}_{(m)}, \mathbfcal{F}_{(m)}, \mathbfcal{H}_{(m)}$, and $\mathbfcal{T}_{(m)}$, respectively.
Model (\ref{eqn30}) is solved by first computing the spatial and temporal graphs online, For this purpose, each matrix $\mathbfcal{X}_{(m)}$ column is used to create the spatial graph $\textbf{G}_{2}^{j}$ in an online manner.
However, single-column data is insufficient to maintain the spatial context; for this reason, we use the final few columns of $\mathbfcal{X}_{(m)}$.
We divide each $\textbf{x}_{m}^{i}$ into non-overlapping patches, much like the batch optimization, and then compute $\mathbfcal{L}_{s(m)}^{i}$.
Every time a new column $\textbf{x}_{m}^{i+1}$ is received, the estimated $\mathbfcal{L}_{s(m)}^{i}$ is updated by discarding the information from the first column $\textbf{x}_{m}^{i}$ and adding the new column information to get $\mathbfcal{L}_{s(m)}^{i+1}$ using Eq. (\ref{eqn7}). 
Similar to this, a temporal graph-based Laplacian matrix $\textbf{L}_{t}^{i+1}$ is estimated by updating the matrix $\textbf{L}_{t}^{i}$ for each new incoming column $\textbf{x}_{m}^{i+1}$.

\subsubsection{Initialization of $\mathbfcal{U}_{(m)}$} By taking the first $r$-column data from matrix $\mathbfcal{X}_{(m)}$, we initialize the basis matrix $\mathbfcal{U}_{(m)}$, and we encode the spatial and temporal information in matrices $\mathbfcal{L}_{s(m)}^{i}$ and $\textbf{L}_{t}^{i}$ as: $\mathbfcal{U}_{(m)}=[\mathbf{\tilde{L}}_{t}^{i}(\textbf{x}_{m}^{1}, \textbf{x}_{m}^{2},\cdot\cdot\cdot,\textbf{x}_{m}^{r})\mathbfcal{L}_{s(m)}^{i}]$, where $\mathbf{\tilde{L}}_{t}^{i}$ is a block of the matrix $\textbf{L}_{t}^{i}$ with the dimensions $r \times r$.
Using this step, $\mathbfcal{U}_{(m)}$ creates a tiny basis matrix that uses little memory.

\subsubsection{Solution for $\textbf{v}_{m}^{i}$} By fixing other variables in Eq. (\ref{eqn30}), a solution to the problem $\textbf{v}_{m}^{i}$ is then formulated as follows:

\begin{equation}
\min_{\textbf{v}_{m}^{i}}||\textbf{x}_{m}^{i}-\mathbfcal{U}_{(m)}\textbf{v}_{m}^{i}-\textbf{f}_{m}^{i}||_{F}^{2} + \frac{1}{2}||\textbf{v}_{m}^{i}||_{F}^{2}.
\label{eqn31}
\end{equation}

\noindent which is solved using a least-square estimation by taking its derivative.
A closed-form solution is then obtained as:

\begin{equation}
\textbf{v}_{m}^{i}=(\mathbfcal{U}_{(m)}^{\top}\mathbfcal{U}_{(m)}+\lambda_{2}\textbf{I})^{-1}\mathbfcal{U}_{(m)}^{\top} (\textbf{x}_{m}^{i}-\textbf{f}_{m}^{i}).
\label{eqn32}
\end{equation}

\subsubsection{Solution for $\textbf{h}_{m}^{i}$} Keeping other variables fixed in (\ref{eqn30}),  a solution to the sub-problem $\textbf{h}_{m}^{i}$ is formulated as follows:

\begin{equation}
\begin{split}
\min_{\textbf{h}_{m}^{i}}||\textbf{h}_{m}^{i}-\textbf{f}_{m}^{i}||_{F}^{2}+ \gamma_{1}\textrm{Tr}(\textbf{h}^{i\top}_{m}\mathbfcal{L}_{s(m)}^{i}\textbf{h}_{m}^{i}),
\end{split}
\label{eqn33}
\end{equation}

\noindent by taking a derivative, we get a closed-form solution of $\textbf{h}_{m}^{i}$ as $\textbf{h}_{m}^{i}=\textbf{f}_{m}^{i} (\gamma_{1}\mathbfcal{L}_{s(m)}^{i}+\textbf{I})^{-1}$, where $\mathbfcal{L}_{s(m)}^{i\top}=\mathbfcal{L}_{s(m)}^{i}$ as it is symmetric.




\subsubsection{Solution for $\textbf{t}_{m}^{i}$} Similar to $\textbf{h}_{m}^{i}$, $\textbf{t}_{m}^{i}$ closed-form solution is obtained as: $\textbf{t}_{m}^{i}=\textbf{f}_{m}^{i} (\gamma_{2}\mathbf{L}_{t}^{i}+\textbf{I})^{-1}$.


\subsubsection{Solution for $\textbf{f}_{m}^{i}$} A solution to this sub-problem is formulated from Eq. (\ref{eqn30}) as:

\begin{equation}
\begin{split}
&{}\min_{\textbf{f}_{m}^{i}}||\textbf{x}_{m}^{i}-\mathbfcal{U}_{(m)}\textbf{v}_{m}^{i}-\textbf{f}_{m}^{i}||_{F}^{2} + ||\textbf{h}_{m}^{i}-\textbf{f}_{m}^{i}||_{F}^{2}+ \lambda||\textbf{f}_{m}^{i}||_{1} \\
&{}+||\textbf{t}_{m}^{i}-\textbf{f}_{m}^{i}||_{F}^{2} =\min_{\textbf{f}_{m}^{i}}\lambda||\textbf{f}_{m}^{i}||_{1}+||\textbf{f}_{m}^{i}-\textbf{q}_{m}^{i}||_{F}^{2}, \textrm{where} \\
&{}\textbf{q}_{m}^{i}=(\textbf{x}_{m}^{i}-\mathbfcal{U}_{(m)}\textbf{v}_{m}^{i}+\textbf{h}_{m}^{i}+\textbf{t}_{m}^{i}) /2.
\end{split}
\label{eqn34}
\end{equation}

\noindent then, a closed-form solution can be obtained using a soft-thresholding operation as $\textbf{f}_{m}^{i}=\textrm{T}_{\lambda}(\textbf{x}_{m}^{i}-\mathbfcal{U}_{(m)}\textbf{v}_{m}^{i}+\textbf{q}_{m}^{i})$.


\subsubsection{Basis $\mathbfcal{U}_{(m)}$ Update} The basis matrix can be updated in two different ways including directly obtaining the closed-form solution and adopting the stochastic gradient descent method.
Using a closed-form solution, we first define two auxilliary matrices, $\textbf{V}^{i \top}_{m}=[\textbf{v}^{1}_{m},\textbf{v}^{2}_{m},\cdot\cdot\cdot,\textbf{v}^{i}_{m}] \in \mathbb{R}^{i \times r}$ and $\textbf{R}^{i}_{m}=[\textbf{r}^{1}_{m},\textbf{r}^{2}_{m},\cdot\cdot\cdot,\textbf{r}^{i}_{m}] \in \mathbb{R}^{p \times i}$, where each $\textbf{r}^{i}_{m}=\textbf{x}_{m}^{i}-\textbf{f}_{m}^{i}$.
$\mathbfcal{U}_{(m)}$ is then updated as follows: $\mathbfcal{U}_{(m)}^{i}=\Theta_{(m)}^{i}(\theta_{(m)}^{i}+\lambda \textbf{I})^{\top}$, where $\Theta_{(m)}^{i}=\Theta_{(m)}^{i-1}+\textbf{r}^{i}_{m}\textbf{v}^{i \top}_{m}$ and $\theta_{(m)}^{i}= \theta_{(m)}^{i-1}+ \textbf{v}^{i}_{m}\textbf{v}^{i\top}_{m}$,
where $\Theta_{(m)}^{i}\in \mathbb{R}^{p \times r}$ and $\theta_{(m)}^{i}\in \mathbb{R}^{r \times r}$.
Using stochastic gradient descent, solution for $\mathbfcal{U}_{(m)}^{i}$ is given by:

\begin{equation}
\begin{split}
{}&\nabla_{\mathbfcal{U}_{(m)}^{i}}f(\mathbfcal{U}_{(m)})= \mathbfcal{U}_{(m)}\textbf{v}^{i}_{m}\textbf{v}^{i\top}_{m}-\textbf{r}^{i}_{m}\textbf{v}^{i\top}_{m}+\lambda\mathbfcal{U}_{(m)} \\
{}& \mathbfcal{U}_{(m)}^{i}\leftarrow\mathbfcal{U}_{(m)}^{i-1}-\eta \nabla_{\mathbfcal{U}_{(m)}^{i}}f(\mathbfcal{U}_{(m)}),
\end{split}
\label{eqn35}
\end{equation}

\noindent where $\eta >0$ is the learning rate.
Then, using average pooling on mode-m foldings, the low-rank $\mathbfcal{B}$ and sparse $\mathbfcal{F}$ tensors are produced as follows:

\begin{equation}
\begin{split}
\mathbfcal{B}= \frac{1}{M} \sum_{m=1}^{M} fold(\mathbfcal{B}_{m}),~\textrm{and}~\mathbfcal{F}= \frac{1}{M} \sum_{m=1}^{M} fold(\mathbfcal{F}_{m}). 
\end{split}
\label{eqn36}
\end{equation}

\noindent Model (\ref{eqn30}) converges to the optimal solution for each instance per-time instance if $\frac{\max (||\textbf{f}^{i}_{m}||_{2},||\textbf{v}^{i}_{m}||_{2})}{p}<\omega$, where $\omega$ is a tolerance parameter for convergence criteria according to \cite{feng2013online, mardani2015subspace}.
Algorithm \ref{algo2} summarizes the O-STRPCA model.

\begin{algorithm}[t!]
	\SetAlgoLined
	\KwIn{$\mathbfcal{X}$, $\gamma_{1} >0$, $\gamma_{2}>0$, $\lambda$, set entries of $\mathbfcal{B}_{m}$, $\mathbfcal{F}_{m}$, $\mathbfcal{L}_{s(m)}^{i}$ \& $\textbf{L}_{t}^{i}$ per-time instance.}
	\textbf{Initialize:} $r>0$, $\omega>0$, $\eta>0$, $\mathbfcal{U}_{(m)}=0$.\\
	\While{$\textrm{not converged}~(k=0,1,..)$}{
         1. Access each column $\textbf{x}_{m}^{i}$ from $\mathbfcal{X}_{m}$. \\
		2. Estimate $\mathbfcal{L}_{s(m)}^{i}$ \& $\textbf{L}_{t}^{i}$ per-time instance.\\  
         3. Estimate $\mathbfcal{U}_{(m)}$.   \\
		4. Estimate $\textbf{v}_{m}^{i}, \textbf{h}_{m}^{i}, \textbf{t}_{m}^{i}$, \& $\textbf{f}_{m}^{i}$ using (\ref{eqn31})-(\ref{eqn34}). \\
         5. Update $\mathbfcal{U}_{(m)}$ using (\ref{eqn35}). \\
         6. Estimate  $\mathbfcal{B}_{(m)}\leftarrow \mathbfcal{U}_{(m)}\mathbfcal{V}_{(m)}$. \\
		7. Check convergence. \\  
			}
	\KwOut{Apply average pooling (\ref{eqn36}) to get $\mathbfcal{B}$ and $\mathbfcal{F}$.}
\caption{Pseudo-code of O-STRPCA model.} 
	\label{algo2}
\end{algorithm}

\section{Experimental Evaluations}
\label{sec:results}
In order to evaluate the effectiveness of the proposed algorithms, we conducted thorough experimental evaluations in this section. 
In the following subsections, we initially discuss the experimental settings and implementation details, benchmark datasets, and performance evaluation metrics followed by the ablation studies before comparing the qualitative and quantitative results of the proposed algorithms with current SOTA approaches. 

\subsection{Experimental Settings and Implementation Details}
We needed some parameters to solve STRPCA (\ref{eqn3}), including $\lambda$, $\gamma_{1}$, $\gamma_{2}$, $k$, $\mu^{0}$, $\zeta$, $\rho$, and $\mu_{max}$.
As recommended by the \cite{lu2016tensor, lu2019tensor}, we set $\lambda=1 / \sqrt{\max (w,h,n)}$.
For both STRPCA and O-STRPCA models, the graph parameters $\gamma_{1}=0.9$ and $\gamma_{2}=1.5$ are established in accordance with the ablation study shown in sec. \ref{sec:ablation}
For the purpose of creating both spatial and temporal graphs, we chose $k=10$ nearest neighbors.
For the spatial graph, we employed a patch size of $a \times a= 8 \times 8$ pixels. 
The other optimization parameters were chosen based on recommendations from \cite{lu2016tensor, lu2019tensor}.
We utilized $r=10$ columns from $\mathbfcal{X}_{m}$ and a learning rate of $\eta=0.01$ in (\ref{eqn35}) to solve the O-STRPCA model (\ref{eqn30}).

All the experiments are carried out on a standard desktop workstation with 128 GB RAM and CPU Intel Xeon E5-2698 V4 2.2 Gz (20-cores) processor.
We implemented both optimization models (\ref{eqn3}) and (\ref{eqn30}) using MATLAB 2023a and LRS Library \footnote{https://github.com/andrewssobral/lrslibrary}. 
We also used a FLANN library \cite{muja2014scalable} for the construction of spatial and temporal graphs.
Furthermore, all results are obtained directly from the author's publications for reasons of comparison, and some of the current RPCA and TRPCA-based background subtraction methods are implemented using the official codes supplied by the authors.

\subsection{Datasets}
\subsubsection{Change Detection 2014 Dataset}
One of the largest background subtraction benchmark datasets, Change Detection 2014 (CD14), has 53 challenging video sequences that were recorded in both indoor and outdoor situations utilizing PTZ, IP, and infrared cameras.
This dataset's sequences are categorized into 11 different categories based on the background scene, including Low Frame Rate (LFR), turbulence, Dynamic Background (DB), Intermittent Object Motion (IOM), Camera Jitter (CJ), shadow, thermal, baseline, Night Videos (NVs), Bad Weather (BW), and PTZ.
There are four to six video sequences in each category.
There are 159,278 video frames in all, with each sequence having pixels that range in size from $320 \times 240$ to $720 \times 576$.
To assess the effectiveness of the background subtraction techniques, ground-truth images of the foreground mask are also given for each sequence.

\subsubsection{I2R Dataset} The  Institute for Infocomm Research (I2R) dataset contains nine video sequences of varying sizes and resolutions captured from both indoor and outdoor scenes using a static camera \cite{li2004statistical}.
The sequences in this dataset experience intensely complicated background alterations, such as crowded foregrounds, gradual and sudden background variations such as flickering of water surface and fountains, etc.
There are a total of 21,901 video frames with the size of each sequence ranging from $120 \times 160$ to $256 \times 320$ pixels.
Additionally, each sequence includes 20 foreground mask ground truth images to assess how well the background subtraction techniques perform.

\subsubsection{BMC 2012 Dataset} 
The Background Models Challenge 2012 (BMC12) dataset contains both synthetic and natural real-time surveillance videos \cite{vacavant2013benchmark}.
Real-time contains nine long-term video sequences that are captured in outdoor scenes and mainly contain the background variation challenges such as the presence of vegetation, cast shadows, the presence of a continuous car flow, weather-changing conditions, sudden lighting conditions, and the presence of big objects.
There are a total of 265715 video frames with the size of each sequence being $240 \times 320$ pixels.
To assess the effectiveness of the background subtraction techniques, several ground-truth images of the foreground mask are also given for each video.

\subsubsection{Wallflower Dataset} The Wallflower dataset contains eight complex video sequences with two sequences overlapped with the I2R dataset.
The videos are captured in outdoor and indoor scenes using a static camera and contain dynamic backgrounds such as waving trees and foreground aperture challenges.
There are a total of 11,748 frames with the size of each sequence being $120 \times 160$ pixels.
To assess the background subtraction methods, one ground-truth image of the foreground mask is provided for each sequence.
\subsubsection{SABS Dataset}
The Stuttgart Artificial Background Subtraction (SABS) dataset contains nine synthetic videos for pixel-wise evaluation of the moving object segmentation \cite{brutzer2011evaluation}.
The dataset introduces common surveillance challenges in the background scene including bootstrap, dynamic background, light switch, darkening, shadow, camouflage, and noisy night.
There are a total of 7,000 frames with the size of each sequence being $600 \times 800$ pixels with several ground truth images.
For a fair comparison of the background subtraction techniques, the sequences in this dataset are separated into training and testing splits.
In order to report on the performance of the proposed algorithms, we solely employed testing sequences.
\subsubsection{SBM-RGBD Dataset}
The Scene Background Modeling RGB Depth (SBM-RGBD) dataset is designed to evaluate the moving object segmentation approaches on both RGB and depth channels \cite{camplani2017benchmarking}.
The dataset contains 33 video sequences captured in indoor scenes which are divided into seven distinct attributes or challenges including illumination changes (strong and mild variations), color camouflage, depth camouflage, intermittent motion, out-of-sensor range, shadows, and bootstrapping.
There are 15,041 frames in all, each having a spatial resolution of $640 \times 480$ pixels and a number of ground-truth images.
The depth channel of each sequence is recorded at either 16 or 8 bits.
To analyze the performance of background subtraction, we employed both RGB and depth channels.

\subsection{Performance Evaluation Metrics}
The $F$-measure score, precision, and recall assessment metrics are used to report the background subtraction performance.
The precision and recall measures serve as the foundation for the $F$-measure score, which may be calculated using: 

\begin{equation}
\begin{split}
    F=\frac{2*Precision*Recall}{Precision + Recall}, \\
    Recall=\frac{TP}{TP+FN},~\textrm{and}~Precision=\frac{TP}{TP+FP},
    \end{split}
\end{equation}

\noindent the true positive, false negative, and false positive, respectively, are denoted by TP, FN, and FP.
The projected background subtraction masks TP, FP, and FN values represent the proportion of moving object pixels that were correctly identified, erroneously categorized, and background pixels, respectively.
High values of these metrics reflect the highest performance of the background subtraction approach.

\subsection{Ablation Studies}
\label{sec:ablation}
\subsubsection{\textbf{Graph-based Hyper-parameters $\gamma_{1}$ and $\gamma_{2}$}} 
\begin{figure}[t!]
\centering
\includegraphics[width=\linewidth,height=2.5in]{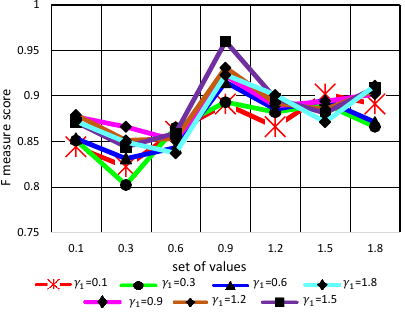}
\caption{Ablation study on setting the hyper-parameters $\gamma_{1}$ and $\gamma_{2}$. 
Averaged $F$-measure score is reported on the nine sequences of the I2R dataset by fixing $\gamma_{1}$ and altering $\gamma_{2}$.
We empirically discovered that the optimum $F$-measure score of 0.960 is attained by setting $\gamma_{1}=0.9$ and $\gamma_{2}=1.5$.}
\label{fig3}
\end{figure}

The ideal graph-based parameters are chosen in this ablation study. 
The model's (\ref{eqn3}) relative relevance is determined by the spatial and temporal graph-based regularization parameters.
Running the background subtraction experiments on the nine sequences from the I2R dataset allowed us to establish these parameters. 
For each parameter, we established a range of values, $\Pi=\{0.1,0.3,0.6,0.9,1.2,1.5,1.8 \}$.
We do tests on these values by setting one regularization parameter $\gamma_{1}$ and altering $\gamma_{2}$ in terms of $\Pi$. 
As a result, we were able to generate $7 \times 7$ combinations of $F$ scores for a single sequence.
By setting one of the two parameters, we tested the STRPCA algorithm on the nine sequences that produced a $7 \times 7 \times 9$ combination.
We calculated the average $7 \times 7$ $F$ scores and presented the results in Fig. \ref{fig3}.
We empirically found that the $F=0.960$ score performs best when $\gamma_{1}=0.9$ and $\gamma_{2}=1.5$ are used.
Then, without making any adjustments, the same set of these parameters is applied to different datasets.

\subsubsection{\textbf{Variants of the Proposed Algorithm}}
We tested different variants of the proposed STRPCA algorithm in this experiment.
We obtain a Temporally-regularized TRPCA (T-TRPCA) model by substituting $\gamma_{1}=0$ and $\gamma_{2}>0$ in the model (\ref{eqn3}).
A Spatially-regularized TRPCA (S-TRPCA) model is obtained by replacing  $\gamma_{2}=0$ and $\gamma_{1}>0$ in (\ref{eqn3}).
In addition, we obtain a classical TRPCA model \cite{lu2019tensor} by replacing $\gamma_{1}= \gamma_{2}=0$ in (\ref{eqn3}).
In order to assess how well integrating spatiotemporal constraints on tensors works, we also implemented a matrix-based form of STRPCA, known as SRPCA.
For this, the RPCA model \cite{wright2009robust} is modified to include the graph-based regularizations presented in Sec. \ref{sec:constraints}, which are then solved using the batch-based ADMM technique of optimization.

Table \ref{table2} displays the effectiveness of all these variations on the Wallflower, SABS, and BMC12 datasets.
Overall, the proposed STRPCA model exhibited the best performance, while the traditional TRPCA model demonstrated significantly degraded performance.
The proposed method significantly outperformed these alternatives by a significant margin, highlighting the advantages of using both spatial and temporal regularizations in the traditional TRPCA model.
The S-TRPCA model outperformed the T-TRPCA model in the SABS dataset, where many sequences exhibit local variations like dark scenes in a background model.
This is due to performance loss caused by the local changes interfering with the temporally similar pixels in a temporal graph.
Numerous sequences in the BMC12 and Wallflower datasets experience global dynamic background fluctuations.
The T-TRPCA model performed better than the S-TRPCA model because the temporal regularization more effectively captured background changes in the temporal domain, whereas the S-TRPCA variant caused many spatially near pixels to become unconnected in the spatial graph.
The STRPCA model demonstrated improved performance on all datasets for the matrix-based SRPCA version, highlighting the advantages of leveraging multi-dimensional tensor data on graph-based constraints.

\begin{table}[t!]
\caption{Performance comparison of different variants of the proposed algorithm including TRPCA, T-TRPCA, and S-TRPCA.
The $F$ measure score for BMC12, Wallflower, and SABS datasets is presented.
The two top-performing algorithms are displayed in red and blue, respectively.}
\begin{center}
\makebox[\linewidth]{
\begin{tabu}{|c|c|c|c|}
\tabucline[0.5pt]{-}
Variants&SABS&BMC12&Wallflower\\\tabucline[0.5pt]{-}
STRPCA&\textcolor{red}{\textbf{0.912}}&\textcolor{red}{\textbf{0.892}}&\textcolor{red}{\textbf{0.951}}\\\tabucline[0.5pt]{-}
SRPCA&0.855&0.816&0.877\\\tabucline[0.5pt]{-}
TRPCA&0.802&0.746&0.792\\\tabucline[0.5pt]{-}
S-TRPCA&\textcolor{blue}{\textbf{0.860}}&0.827&0.837\\\tabucline[0.5pt]{-}
T-TRPCA&0.844&\textcolor{blue}{\textbf{0.856}}&\textcolor{blue}{\textbf{0.905}}\\\tabucline[0.5pt]{-}
\end{tabu}
}
\end{center}
\label{table2}
\end{table}

\subsubsection{\textbf{Nearest neighbor and Patch size in Spatial Graph}}
In order to build spatial and temporal graphs, we also carried out an ablation study on various nearest neighbor ($k$) and patch size ($a \times a$) values.
The $F$ score for the I2R dataset is shown in Table \ref{table3} by altering the values of $k$ and $a \times a$.
On this dataset, choosing $k=10$ and a patch size of $8 \times 8$ yields the best results.

\begin{table}[t!]
\caption{Performance comparison of the proposed STRPCA algorithm on the I2R dataset for various $k$ and patch sizes $a \times a$ values.
The best performance is of 0.892 is observed using $k=10$ and $a \times a=8 \times 8$.}
\begin{center}
\makebox[\linewidth]{
\scalebox{0.92}{
\begin{tabu}{|c|c|c|c|c|c|}
\tabucline[0.5pt]{-}
$k$&$k=2$&$k=4$&$k=6$&$k=8$&$k=10$\\\tabucline[0.5pt]{-}
STRPCA&0.871&0.896&0.902&0.941&\textbf{0.960}\\\tabucline[0.5pt]{-}
$a \times a$&$4 \times 4$&$8 \times 8$&$12 \times 12$&$16 \times 16$&$20 \times 20$\\\tabucline[0.5pt]{-}
STRPCA&0.935&\textbf{0.960}&0.907&0.891&0.875\\\tabucline[0.5pt]{-}
\end{tabu}
}}
\end{center}
\label{table3}
\end{table}

\subsection{Comparison with SOTA Methods}
We compared the results of our proposed algorithms with 15 SOTA methods on six publicly available benchmark datasets.
We selected three categories of the SOTA methods including RPCA-based, TRPCA-based, and deep learning for background subtraction.

The five RPCA-based  methods included classical RPCA \cite{wright2009robust}, MSCL \cite{javed2017background}, TS-RPCA \cite{ebadi2017foreground}, LSD \cite{liu2015background}, and OMoGMF$+$TV \cite{yong2017robust}.
The classical RPCA model's organized sparsity was enhanced by these approaches.
MSCL method regularizes subspace clustering constraints, TS-RPCA encodes dynamic tree-structured constraints, LSD incorporates a group sparsity structure in the form of overlapped pixels within the spare component, and OMoGMF$+$TV method is based on a mixture of Gaussian distribution, which is updated online frame by frame in an online manner.
The five TRPCA methods included classical TRPCA \cite{lu2019tensor}, ORLTM \cite{li2018online}, NIOTenRPCA \cite{li2022tensor}, TV-TRPCA \cite{cao2016total}, and ETRPCA \cite{gao2020enhanced}.
ORLTM is an online tensor-based method that incorporates a background dictionary.
NIOTenRPCA is based on online compressive video reconstruction and background subtraction that explicitly models the background disturbance.
TV-RPCA addresses the problem of structured sparsity of the tensor by explicitly modeling the total variation norm.
ETRPCA explicitly considers the salient difference information of the input pixels between singular values of tensor data by the weighted tensor Schatten p-norm minimization.

The five deep learning-based methods included ZBS \cite{an2023zbs}, 3DCD \cite{mandal20203dcd}, CrossNet \cite{liang2023crossnet}, CascadeCNN \cite{wang2017interactive}, and STPNet \cite{yang2021stpnet}.
ZBS is an unsupervised deep learning technique that relies on zero-shot object detection.
3DCD is a fully supervised method that exploits scene-independent and scene-dependent evaluations to test the supervised methods in completely unseen videos for generalized background subtraction tasks.
CrossNet employs an end-to-end cross-scene network via 3D optical flow information to address scene-specific challenges.
CascadeCNN is a fully supervised block-based method.
STPNet is an end-to-end propagation network that captures spatial and temporal features, simultaneously.
All these existing deep learning methods train their loss functions on the training sequences and perform evaluations on the testing sequences.
Therefore, these methods are data-hungry and totally rely on manual labeling as well as supervised training for differentiating background-foreground pixels.
On the other hand, subspace learning methods such as RPCA and TRPCA etc., are completely unsupervised methods that do not employ any labels for background subtraction.

\begin{figure*}[t!]
\centering
\includegraphics[width=\linewidth]{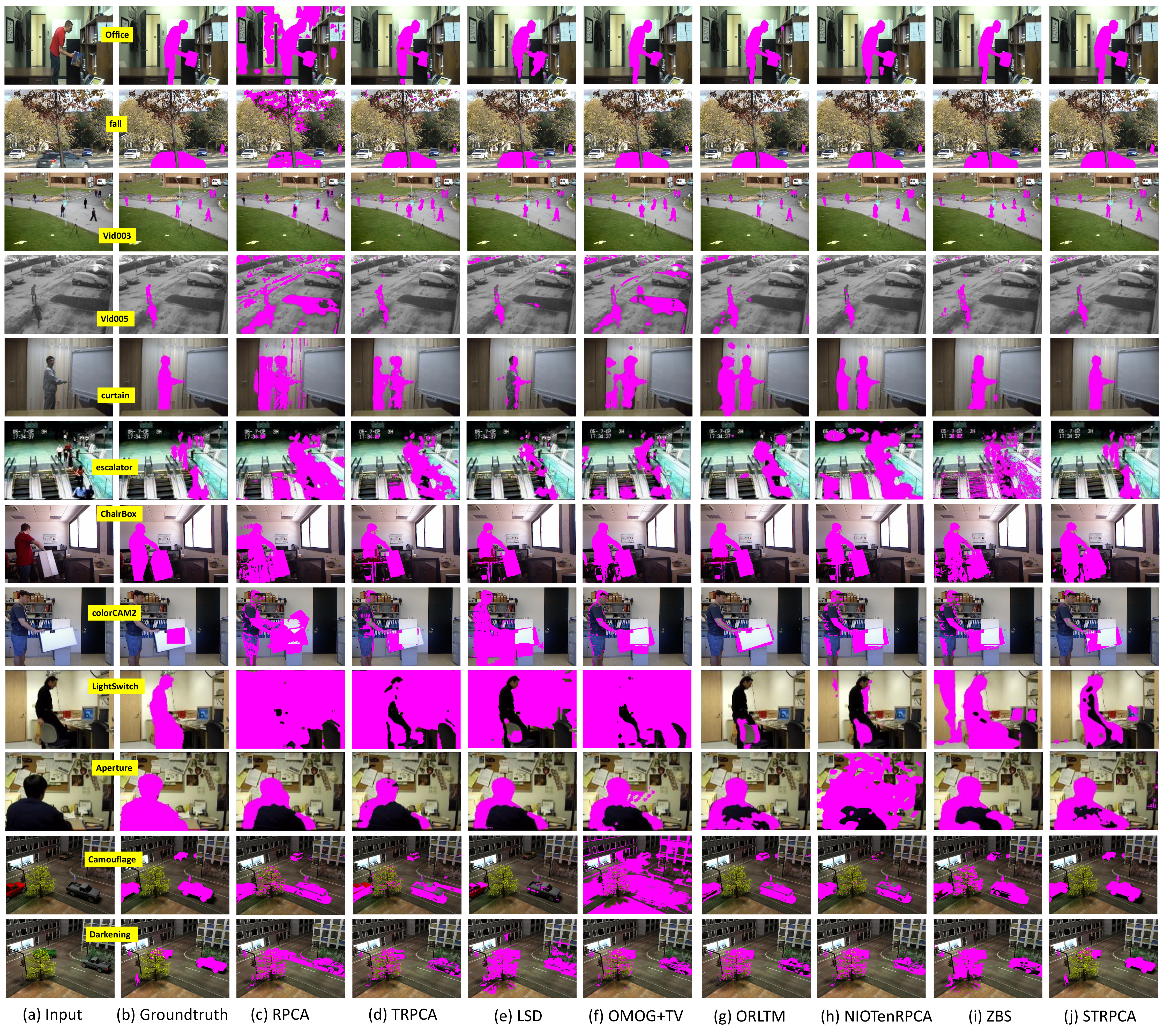}
\caption{Comparison of 12 sequences chosen from each dataset for background subtraction using published techniques.
Input images, ground truth images, and background subtraction estimates from RPCA \cite{wright2009robust}, TRPCA \cite{lu2019tensor}, LSD \cite{liu2015background}, OMOG$+$TV \cite{yong2017robust}, ORLTM \cite{li2018online}, NIOTenRPCA \cite{li2022tensor}, ZBS \cite{an2023zbs}, and the proposed STRPCA method are shown from left to right.
When compared to the SOTA approaches, the STRPCA algorithm produces superior visual results.}
\label{fig4}
\end{figure*}

\subsection{Qualitative Evaluations}
The visual results of the proposed STRPCA algorithm on 12 difficult sequences chosen from the aforementioned six datasets are shown in Fig. \ref{fig4}, along with a comparison to published approaches. 
On every sequence depicted in Fig. \ref{fig4} from top to bottom, STRPCA beat the SOTA approaches. 
This is because graph-based spatiotemporal constraints were included and produced foregrounds with complete spatial structures.

In particular, RPCA (Fig. \ref{fig4} (c)) showed a considerable performance loss because it was unable to manage the local fluctuations and swaying of the bushes problems of CD12 sequences (\textit{office} and \textit{fall}).
In these sequences, the majority of the compared approaches did a superior job of background-foreground segregation.
The BMC12 sequences, particularly \textit{Vid003} and \textit{Vid005}, had challenges of static backgrounds and bad weather conditions.
On \textit{Vid003}, all of the approaches under comparison—aside from RPCA—displayed too smoothed foreground segmentation (Figs. \ref{fig4} (d)-(i)); however, on \textit{Vid005}, certain methods, such as TRPCA, LSD, ORLTM, and ZBS, delivered better outcomes than the STRPCA.
The dynamic background and bootstrapping difficulties are demonstrated in the I2R dataset's (\textit{curtain} and \textit{escalator} sequences).
Only the STRPCA produced superior results; the other approaches could not handle the problems better because of ghost artifacts and too-smoothed foreground segmentation in the background scene.
The comparative approaches also failed to give valid results for the remaining sequences of the SBM-RGBD, Wallflower, and SABS datasets because these sequences had problems with camouflage, light change, aperture, and dynamic background modeling. 

Overall, STRPCA showed superior qualitative results in comparison to the current subspace learning techniques, highlighting the advantages of taking into account graph-based spatiotemporal constraints inside the sparse component.

\subsection{Quantitative Evaluations}
\subsubsection{\textbf{Evaluations on CD14 Dataset}}
We compared the performance of the proposed algorithms with two distinct paradigms, including deep learning-based techniques and subspace learning, on this dataset. 
While deep learning-based approaches are fully supervised methods that depend on training data, subspace learning methods like RPCA and TRPCA are fully unsupervised methods.\\
\noindent \textbf{Comparisons with RPCA and TRPCA-based Methods:} 
On the CD14 dataset, Table \ref{table4} compares the performance in terms of average $F$ scores of the proposed algorithms overall and by category with 10 other RPCA- and TRPCA-based SOTA approaches. 

Overall, the $F$ scores for STRPCA and O-STRPCA, were 89.80$\%$ and 86.20$\%$, respectively, which is much better than the approaches under comparison. 
The CD14 dataset's difficult sequences could not be processed well by the compared approaches, which led to a performance reduction.
STRPCA achieved 8.10$\%$  and 5.10$\%$ greater accuracy when compared to the MSCL and TV-TRPCA approaches, whereas O-STRPCA produced 4.50$\%$ and 1.50$\%$ higher performance. 
Our proposed regularizations inside the batch-based and online-based optimization models have contributed to these improved results.

All of the tested approaches were able to achieve a $F$ score of more than 75.00$\%$ for the baseline sequences (4 videos), showing that these sequences did not significantly challenge the compared methods.  STRPCA achieved the highest $F$ score of 98.10$\%$.
The six sequences in the dynamic background category include difficult scenes with flowing fountains, swinging shrubs, and rippling water surfaces. 
Most of the compared approaches had significant problems with these sequences ($F$ score less than 80.00$\%$). 
STRPCA and O-STRPCA algorithms achieved the best accuracies of 95.50$\%$ and 91.10$\%$, respectively, in this category. 
The best performance was achieved by TS-RPCA due to the tree-structured sparsity constraints and the TV-TRPCA due to the total variation norm.

The difficulty recognized for producing ghosting artifacts in the detected motion, i.e., abandoned foreground objects or deleted foreground objects, is present in the intermittent object motion category (6 videos).
With the exception of MSCL and TV-TRPCA, the bulk of the examined approaches were unable to handle these sequences, whereas our proposed algorithms greatly outperformed them. 
The motionless frames were removed in MSCL using optical flow, and the structured foreground areas were modeled by TV-TRPCA. 
Our proposed algorithms produced lower $F$ scores of 85.00$\%$ and 83.60$\%$ when compared to baseline and dynamic background sequences, respectively. 
The STRPCA and TV-TRPCA methods achieved the best and second-best performances in the turbulence category (4 videos), whereas MSCL showed favorable performance.

Similar to this, only STRPCA and O-STRPCA algorithms were able to achieve $F$ scores of greater than 80.00$\%$ or 90.00$\%$ in other categories such as low frame rate (4 videos), camera jitter (4 videos), thermal (5 videos), night videos (6 videos), and bad weather (4 videos), which contain more difficult background modeling challenges. 
The six videos in the shadow category include scenes with a mix of soft and harsh shadows with sporadic tints. 
The TS-RPCA technique achieved the greatest performance of 91.70$\%$, whereas our proposed algorithm's performance degraded by 6.30$\%$ and 2.50$\%$, respectively, due to their inability to manage shadow in background scenes.\\

\noindent \textbf{Comparisons with Deep Learning-based Methods:} On the training videos of the CD14 dataset, these approaches first learn deep feature representations in an end-to-end fashion, and then they evaluate those representations using either the visible or unseen testing sequences. 
We used the same testing split established by \cite{wang2017interactive} to compare the SOTA approaches fairly, and we assessed our proposed STRPCA algorithm.

Comparing the proposed STRPCA's quantitative performance to that of the available deep learning techniques is shown in Table \ref{table5}.
Overall, it can be shown that the unsupervised approaches including ZBS and our STRPCA are less popular than the fully supervised deep learning methods.
Since STRPCA does not incorporate learning from the training sequences, it nevertheless demonstrated a performance deterioration of 2.60$\%$ percent when compared to the CascadeCNN technique, which had the greatest results. 
STRPCA got 4.70$\%$  and 1.30$\%$ greater $F$ scores than the fully supervised STPNet and unsupervised ZBS approaches, demonstrating its superiority compared to end-to-end training.

\begin{table*}[t!]
\caption{Comparison of the proposed algorithms' performance with published unsupervised RPCA- and TRPCA-based approaches on the 10 categories of the CD14 dataset. 
For each category, the average $F$ measure score is given.
The top and second-best performing approaches are shown, respectively, by the colors red and blue.}
\begin{center}
\makebox[\linewidth]{
\scalebox{0.90}{
\begin{tabu}{|c|c|c|c|c|c|c|c|c|c|c|c|}
\tabucline[2.0pt]{-}
Methods&Baseline&DB&IOM&Shadow&Turbulence&LFR&CJ&Thermal&NVs&BW&Average\\\tabucline[2.0pt]{-}
RPCA&0.751&0.692&0.482&0.735&0.681&0.644&0.621&0.651&0.601&0.711&0.656\\\tabucline[0.5pt]{-}
MSCL&0.874&0.859&0.805&0.825&0.801&0.761&0.833&0.803&0.784&0.831&0.817\\\tabucline[0.5pt]{-}
TS-RPCA&\textbf{\textcolor{blue}{0.943}}&0.905&0.783&\textbf{\textcolor{red}{0.917}}&-&-&0.880&0.719&-&-&-\\\tabucline[0.5pt]{-}
LSD&0.921&0.711&0.677&0.811&0.704&0.741&0.781&0.750&0.771&0.792&0.765\\\tabucline[0.5pt]{-}
OMoGMF$+$TV&0.851&0.762&0.716&0.682&0.611&0.755&0.782&0.704&0.751&0.785&0.739\\\tabucline[0.5pt]{-}
TRPCA&0.781&0.702&0.511&0.744&0.701&0.651&0.654&0.666&0.621&0.722&0.675\\\tabucline[0.5pt]{-}
ORLTM&0.811&0.744&0.788&0.806&0.751&0.774&0.786&0.717&0.702&0.771&0.765\\\tabucline[0.5pt]{-}
NIOTenRPCA&0.795&0.801&0.766&0.808&0.791&0.776&0.791&0.703&0.732&0.762&0.772\\\tabucline[0.5pt]{-}
TV-TRPCA&0.855&0.891&0.822&0.866&\textbf{\textcolor{blue}{0.841}}&0.791&0.891&0.851&0.805&0.866&0.847\\\tabucline[0.5pt]{-}
ETRPCA&0.792&0.731&0.568&0.781&0.721&0.681&0.682&0.681&0.651&0.735&0.702\\\tabucline[2.0pt]{-}
Ours (O-STRPCA)&0.931&\textbf{\textcolor{blue}{0.911}}&\textbf{\textcolor{red}{0.850}}&\textbf{\textcolor{blue}{0.854}}&0.833&\textbf{\textcolor{blue}{0.821}}&\textbf{\textcolor{blue}{0.912}}&\textbf{\textcolor{blue}{0.861}}&\textbf{\textcolor{blue}{0.822}}&\textbf{\textcolor{blue}{0.881}}&\textbf{\textcolor{blue}{0.862}}\\\tabucline[0.5pt]{-}
Ours (STRPCA)&\textbf{\textcolor{red}{0.981}}&\textbf{\textcolor{red}{0.955}}&\textbf{\textcolor{blue}{0.836}}&\textbf{\textcolor{blue}{0.892}}&\textbf{\textcolor{red}{0.871}}&\textbf{\textcolor{red}{0.842}}&\textbf{\textcolor{red}{0.944}}&\textbf{\textcolor{red}{0.895}}&\textbf{\textcolor{red}{0.853}}&\textbf{\textcolor{red}{0.902}}&\textbf{\textcolor{red}{0.898}}\\\tabucline[2.0pt]{-}
\end{tabu}
}}
\end{center}
\label{table4}
\end{table*}

\begin{table*}[t!]
\caption{On 10 categories from the CD14 dataset, the proposed UnSupervised (US) STRPCA algorithm was compared to published Fully Supervised (FS) approaches based on deep learning. 
The average $F$ measure score for each category is presented. 
The top and second-best performing approaches are shown, respectively, by the colors red and blue. 
Overall, STRPCA shows favorable performance compared to FS deep learning-based methods.}
\begin{center}
\makebox[\linewidth]{
\scalebox{0.90}{
\begin{tabu}{|c|c|c|c|c|c|c|c|c|c|c|c|}
\tabucline[2.0pt]{-}
Methods&Baseline&DB&IOM&Shadow&Turbulence&LFR&CJ&Thermal&NVs&BW&Average\\\tabucline[2.0pt]{-}
ZBS (US)&\textbf{\textcolor{red}{0.965}}&\textbf{\textcolor{blue}{0.929}}&0.875&\textbf{\textcolor{red}{0.976}}&0.635&0.743&\textbf{\textcolor{red}{0.954}}&\textbf{\textcolor{blue}{0.869}}&0.680&\textbf{\textcolor{blue}{0.922}}&0.851\\\tabucline[0.5pt]{-}
3DCD (FS)&0.930&0.870&\textbf{\textcolor{red}{0.900}}&0.890&\textbf{\textcolor{red}{0.940}}&0.760&0.830&\textbf{\textcolor{red}{0.870}}&\textbf{\textcolor{blue}{0.860}}&\textbf{\textcolor{red}{0.940}}&\textbf{\textcolor{blue}{0.879}}\\\tabucline[0.5pt]{-}
CrossNet (FS)&0.933&0.901&\textbf{\textcolor{blue}{0.881}}&0.861&\textbf{\textcolor{blue}{0.902}}&0.701&0.912&0.855&0.799&0.908&0.865\\\tabucline[0.5pt]{-}
CascadeCNN (FS)&0.951&\textbf{\textcolor{red}{0.942}}&0.812&\textbf{\textcolor{blue}{0.922}}&0.871&\textbf{\textcolor{red}{0.802}}&\textbf{\textcolor{blue}{0.944}}&0.866&\textbf{\textcolor{red}{0.871}}&\textbf{\textcolor{blue}{0.922}}&\textbf{\textcolor{red}{0.890}}\\\tabucline[0.5pt]{-}
STPNet (FS) &\textbf{\textcolor{blue}{0.958}}&0.805&0.826&0.911&0.724&0.729&0.772&0.868&0.696&0.889&0.817\\\tabucline[2.0pt]{-}
Ours (STRPCA)&0.933&0.902&0.861&0.851&0.867&\textbf{\textcolor{blue}{0.791}}&0.833&0.865&0.844&0.899&0.864\\\tabucline[2.0pt]{-}
\end{tabu}
}}
\end{center}
\label{table5}
\end{table*}

\subsubsection{\textbf{Evaluations on BMC12 Dataset}}The performance comparison of the proposed algorithms with the SOTA RPCA and TRPCA-based techniques is shown in Table \ref{table6}.
Overall, both STRPCA and O-STRPCA perform better on average for the exceedingly difficult natural sequences of the BMC12 dataset. 
This is due to the fact that our algorithms can deal with complex and dynamic backgrounds like continuously moving cars and changing weather. 
With the help of the built-in spatial and temporal graph-based constraints, this capacity enables it to successfully segregate real, well-defined foreground pixels.

\subsubsection{\textbf{Evaluations on Wallflower Dataset}}
The performance comparison between the proposed algorithms and the published approaches on the Wallflower dataset is shown in Table \ref{table6}  in terms of the average $F$ measure score. 
Overall, the tree-structured induced norm in TS-RPCA received the second-highest score (93.30$\%$), whereas the STRPCA method performed the best (96.10$\%$). 
Comparing our online alternative, O-STRPCA, to the TS-RPCA technique, it showed a similar performance of 91.10$\%$. 
The proposed algorithms were successful in handling the dataset's sudden illumination change sequences, which our low-rank tensor only partially made up for.

\subsubsection{\textbf{Evaluations on I2R Dataset}}
The performance comparison using the I2R dataset is also shown in Table \ref{table6}. Overall, STRPCA outperforms in nine sequences, and its $F$ score is around 2.20$\%$ percent higher than that of the second-best TS-RPCA (batch technique). 
More precisely, as compared to the approaches that have been published, both STRPCA and O-STRPCA exhibit more encouraging performance on this dataset. 
The first is that the graph-based constraints model the sparse component better even in the presence of complex background scenes, and the second is that it fully utilizes the underlying information of video sequences by learning the sparsity structure of the data across all tensor modes. 
These two factors are what make our proposed algorithms superior to others in the field.

\subsubsection{\textbf{Evaluations on SBM-RGBD Dataset}}
Table \ref{table6} shows a performance comparison of the proposed algorithms on this dataset. The outcomes show that our proposed algorithms outperformed all unsupervised RPCA and TRPCA-based SOTA methods. 
Particularly, the SOTA MSCL technique was outperformed by STRPCA and O-STRPCA by about 10.00$\%$ and 4.00$\%$, respectively, demonstrating the benefits of the proposed regularization.

\begin{table}[t!]
\caption{Performance comparison of the proposed algorithms on the Wallflower, I2R, BMC12, and SBM-RGBD datasets with established approaches. The average $F$ measure score is reported for each dataset.
The top and second-best performing approaches are shown, respectively, by the colors red and blue.}
\begin{center}
\makebox[\linewidth]{
\scalebox{0.80}{
\begin{tabu}{|c|c|c|c|c|}
\tabucline[2.0pt]{-}
Methods&Wallflower&I2R&BMC12&SBM-RGBD\\\tabucline[2.0pt]{-}
RPCA&0.731&0.728&0.691&0.665\\\tabucline[0.5pt]{-}
MSCL&0.860&0.820&0.820&0.800\\\tabucline[0.5pt]{-}
TS-RPCA&\textbf{\textcolor{blue}{0.933}}&0.933&-&-\\\tabucline[0.5pt]{-}
LSD&0.759&0.752&0.750&0.700\\\tabucline[0.5pt]{-}
OMoGM+TV&0.820&0.770&0.750&0.780\\\tabucline[0.5pt]{-}
TRPCA&0.766&0.735&0.728&0.701\\\tabucline[0.5pt]{-}
ORLTM&0.752&0.788&0.744&0.732\\\tabucline[0.5pt]{-}
NIOTenRPCA&0.781&0.751&0.744&0.699\\\tabucline[0.5pt]{-}
TV-TRPCA&0.788&0.771&0.751&0.731\\\tabucline[0.5pt]{-}
ETRPCA&0.771&0.762&0.735&0.722\\\tabucline[2.0pt]{-}
STRPCA&\textbf{\textcolor{red}{0.961}}&\textbf{\textcolor{red}{0.960}}&\textbf{\textcolor{red}{0.881}}&\textbf{\textcolor{red}{0.901}}\\\tabucline[0.5pt]{-}
O-STRPCA&0.911&\textbf{\textcolor{blue}{0.948}}&\textbf{\textcolor{blue}{0.851}}&\textbf{\textcolor{blue}{0.867}}\\\tabucline[2.0pt]{-}
\end{tabu}
}}
\end{center}
\label{table6}
\end{table}

\subsubsection{\textbf{Evaluations on SABS Dataset}}
STRPCA algorithm was superior in six out of nine video sequences, including Basic, Dynamic Background, Darkening, Camouflage, etc., as shown in the findings in Table \ref{table7}. H.264 Compression, Light Switch, and Bootstrap sequences, on the other hand, were incompatible with our algorithms.
This is due to the STRPCA method's difficulty in adapting to background perturbations, such as the irregular background distribution in these settings, which caused our system to suffer from these problems.
Overall, STRPCA outperformed SOTA approaches by an average of 87.40$\%$, while O-STRPCA came in second place behind the TS-RPCA method.

\begin{table*}[t!]
\caption{Performance comparison of nine sequences from the SABS datasets using the proposed algorithms and published approaches.
Each sequence's average $F$-measure score is given. 
The top and second-best performing approaches are shown, respectively, by the colors red and blue.}
\begin{center}
\makebox[\linewidth]{
\scalebox{0.80}{
\begin{tabu}{|c|c|c|c|c|c|c|c|c|c|c|}
\tabucline[2.0pt]{-}
Methods&Basic&Dynamic&Booststrap&Darkening&Light&Noisy &Camouflage&No &H.264 &Average\\
&&Background&&&Switch&Night&&Camouflage&Compression&\\\tabucline[2.0pt]{-}
RPCA&0.680&0.652&0.702&0.651&0.588&0.622&0.633&0.722&0.751&0.666\\\tabucline[0.5pt]{-}
MSCL&0.820&0.800&\textbf{\textcolor{red}{0.840}}&0.820&\textbf{\textcolor{red}{0.790}}&0.750&0.800&0.790&0.810&0.800\\\tabucline[0.5pt]{-}
TS-TRPCA&\textbf{\textcolor{blue}{0.867}}&0.871&\textbf{\textcolor{blue}{0.822}}&\textbf{\textcolor{blue}{0.907}}&0.570&0.897&\textbf{\textcolor{blue}{0.894}}&0.913&\textbf{\textcolor{red}{0.841}}&0.842\\\tabucline[0.5pt]{-}
LSD&0.730&0.720&0.770&0.700&0.620&0.720&0.750&0.800&0.800&0.720\\\tabucline[0.5pt]{-}
OMoGM+TV&0.800&0.740&0.700&0.680&0.570&0.790&0.750&0.820&0.770&0.730\\\tabucline[0.5pt]{-}
TRPCA&0.711&0.681&0.732&0.681&0.621&0.671&0.756&0.747&0.773&0.708\\\tabucline[0.5pt]{-}
ORLTM&0.733&0.714&0.752&0.718&0.641&0.705&0.776&0.762&0.801&0.733\\\tabucline[0.5pt]{-}
NIOTenRPCA&0.722&0.736&0.771&0.724&0.671&0.781&0.791&0.823&0.827&0.760\\\tabucline[0.5pt]{-}
TV-TRPCA&0.855&0.821&0.798&0.794&\textbf{\textcolor{blue}{0.715}}&0.821&0.854&0.791&0.796&0.805\\\tabucline[0.5pt]{-}
ETRPCA&0.733&0.701&0.756&0.702&0.655&0.701&0.761&0.756&0.781&0.727\\\tabucline[2.0pt]{-}
Ours (STRPCA)&\textbf{\textcolor{red}{0.887}}&\textbf{\textcolor{red}{0.901}}&0.802&\textbf{\textcolor{red}{0.941}}&0.702&\textbf{\textcolor{red}{0.933}}&\textbf{\textcolor{red}{0.921}}&\textbf{\textcolor{red}{0.952}}&\textbf{\textcolor{blue}{0.833}}&\textbf{\textcolor{red}{0.874}}\\\tabucline[0.5pt]{-}
Ours (O-STRPCA)&0.841&\textbf{\textcolor{blue}{0.888}}&0.791&0.902&0.671&\textbf{\textcolor{blue}{0.918}}&0.889&\textbf{\textcolor{blue}{0.923}}&0.802&\textbf{\textcolor{blue}{0.847}}\\\tabucline[2.0pt]{-}
\end{tabu}
}}
\end{center}
\label{table7}
\end{table*}

\subsection{Computational Cost and Running Time Analysis}
The computational complexity and execution time of the proposed algorithms were also evaluated. 
The calculation of the spatiotemporal graphs and the update of the $\mathbfcal{B}$ and $\mathbfcal{F}$ tensors are the two computationally expensive operations for the batch-based technique (STRPCA). 
The $\mathbfcal{X}_{3}$ columns are where the temporal graph is built, hence the cost of computing it is $O(snlog(n))$, where $s=w \times h$ is the number of pixels in each column and $n$ is the total number of columns in $\mathbfcal{X}_{3}$. 
The cost of constructing the spatial graph $\textbf{G}_{2}^{j}$, which is patch-wise among the input tensor's frontal slices $\mathbfcal{X}^{(j)}$, is $O(snlog(s))$.
Similar to this, the update of $\mathbfcal{B}^{k+1}$ in the $k$-th iteration heavily influences the computing cost of optimizing the model (\ref{eqn3}).
The overall cost of updating a $\mathbfcal{B}^{k+1}$ tensor is $O\Big (Tsnlog(n)+Tmax(w,h)min^{2}(w,h)n\Big)$, where $T$ is the total number of iterations. 
This is because updating a $\mathbfcal{B}^{k+1}$ tensor needs an estimation of FFT and $n$ SVDs operation of $w \times h$ matrices.
Last but not least, $O\Big(n [s(log(n))+log(s))+T(max(w,h)min^{2}(w,h))]\Big)$ is the overall computing complexity for solving (\ref{eqn3}) with graph-based regularization.

For the online model (O-STRPCA), estimating the column vectors $\{\textbf{v}_{m}, \textbf{f}_{m}\}$, basis matrix $\mathbfcal{U}_{m}$, and the accumulation matrices $\textbf{V}_{m}$, $\Theta_{m}$, and $\theta_{m}$ determines the computing cost. 
The linear operation needed to compute the vectors $\textbf{v}_{m}$ and $\textbf{f}_{m}$ is $O(slog(s))$. 
Additionally, updating $\mathbfcal{U}_{m}$ costs $O(sr^{2})$ whereas updating $\textbf{V}_{m}$, $\Theta_{m}$, and $\theta_{m}$ matrices costs $O(slog(r))$, where $r$ is a rank in online processing.
As a result, the overall complexity for solving the model (\ref{eqn30}) is given by the expression $O\Big( s [r^{2} + log(s)+ log(r)]\Big)$, which demonstrates that it does not rely on the number of frames and is proportional to $r$. 
This cost is appealing and so overcomes the problems with the TRPCA's real-time processing \cite{lu2019tensor}.

The proposed algorithms' running times are also provided and contrasted with those of published techniques including RPCA, TRPCA, and ORLTM.
We used a frames per second (fps) assessment metric for this purpose, and we then recorded the computing time on a sequence termed \textit{office} that had 2050 frames with $320 \times 240$ spatial resolution that was taken from the CD12 dataset. 
The ORLTM technique took 5.97 fps while RPCA and TRPCA took 3.90 fps and 2.40 fps, respectively. 
Both of our proposed algorithms, STRPCA and O-STRPCA, ran at speeds of 1.60 and 4.30fps.
Compared to TRPCA, the STRPCA method requires more time due to the incorporation of graph-based restrictions. 
Tables \ref{table4}-\ref{table7} demonstrate that both methods work well when it comes to background subtraction, however, our O-STRPCA is faster than its batch equivalent.

\section{Conclusion and Future Work}
\label{sec:conclusion}
We developed novel TRPCA-based algorithms in this study to learn structured sparse tensors for background subtraction tasks. 
We suggested requiring graph-based Laplacian regularizations in both space and time for the traditional TRPCA technique.
We created two graphs one in the time domain and the other in the spatial domain for this aim.
For more reliable background subtraction, the Laplacian matrices generated from these graphs enforced the structural sparsity inside the sparse component. 
We used batch and online optimization techniques to solve the proposed model. 
In contrast to the online optimization algorithm O-STRPCA, which handled video sequences sequentially by using a reformulated form of the nuclear norm constraints, the proposed STRPCA is more efficient and requires that all video frames be loaded into memory in order to handle large datasets. 
Extensive experimental results show that, when compared to various SOTA approaches, our proposed algorithms perform more favorably and show promising outcomes on six publicly accessible background subtraction benchmark datasets. In contrast to SOTA approaches, our algorithms were able to handle complicated background scenes in the presence of lighting conditions, local and global background fluctuations, and camouflage. Training a TRPCA-based deep neural network for background-foreground separation will be the foundation of our future work.

\ifCLASSOPTIONcaptionsoff
  \newpage
\fi

\bibliographystyle{IEEEtranS}
\bibliography{bare_jrnl}
\end{document}